\documentclass[runningheads]{llncs}
\usepackage{graphicx}

\usepackage{tikz}
\usepackage{comment}
\usepackage{amsmath,amssymb} %
\usepackage{color}
\usepackage{wrapfig}
\usepackage{colortbl}
\usepackage{threeparttable}
\usepackage{multirow,mathtools} 
\newcommand{\etal}{\textit{et al.}}
\usepackage[accsupp]{axessibility}  %

\DeclareMathOperator{\Prob}{\mathbb{P}}
\DeclareMathOperator{\E}{\mathbb{E}}
\DeclareMathOperator*{\argmin}{arg\,min}

\begin{document}
\pagestyle{headings}
\mainmatter
\def\ECCVSubNumber{2900}  %

\title{Unified Implicit Neural Stylization} %

\titlerunning{Unified Implicit Neural Stylization} 
\authorrunning{Z. Fan, et al.} 
\author{%
  Zhiwen Fan\inst{1, \star}\index{Zhiwen, Fan}, Yifan Jiang\inst{1, \star}\index{Yifan, Jiang}, Peihao Wang\inst{1, \star}\index{Peihao, Wang},\\ Xinyu Gong\textsuperscript{1}\index{Xinyu, Gong},
  Dejia Xu\textsuperscript{1}\index{Dejia, Xu}, Zhangyang Wang\inst{1}\index{Zhangyang, Wang}\\
}
\institute{\textsuperscript{1}The University of Texas at Austin \\
\email{\{zhiwenfan,yifanjiang97,peihaowang,xinyu.gong,dejia,atlaswang\}@utexas.edu}}

\newcommand{\TODO}[1]{{\color{red}[zw: #1]}}
 
\newcommand{\Mat}{\boldsymbol}
\newcommand{\Set}{\mathcal}
\newcommand{\G}{}
\newcommand{\real}{\mathbb{R}}
\newcommand{\complex}{\mathbb{C}}
\newcommand{\FT}{\mathcal{F}}
\newcommand{\IFT}{\mathcal{F}^{-1}}

% \titlerunning{Abbreviated paper title}
\maketitle

\begin{abstract}
Representing visual signals by implicit neural representation (INR) has prevailed among many vision tasks. 
Its potential for editing/processing given signals remains less explored.
This work explores a new intriguing direction: training a \textbf{stylized} implicit representation, using a generalized approach that can apply to various 2D and 3D scenarios. We conduct a pilot study on a variety of INRs, including 2D coordinate-based representation, signed distance function, and neural radiance field. Our solution is a Unified \textbf{I}mplicit \textbf{N}eural \textbf{S}tylization framework, dubbed \textbf{INS}. In contrary to vanilla INR, INS decouples the ordinary implicit function into a style implicit module and a content implicit module, in order to separately encode the representations from the style image and input scenes. An amalgamation module is then applied to aggregate these information and synthesize the stylized output. To regularize the geometry in 3D scenes, we propose a novel self-distillation geometry consistency loss which preserves the geometry fidelity of the stylized scenes. Comprehensive experiments are conducted on multiple task settings, including fitting images using MLPs, stylization for implicit surfaces and sylized novel view synthesis using neural radiance. We further demonstrate that the learned representation is continuous not only spatially but also style-wise, leading to effortlessly interpolating between different styles and generating images with new mixed styles. Please refer to the video on our project page for more view synthesis results: \url{https://zhiwenfan.github.io/INS}.
\end{abstract}

\section{Introduction}
Implicit Neural Representation (INR) has gained remarkable popularity \let\thefootnote\relax\footnote{$\star$ Equal contribution} in representing concise signal representation in computer vision and computer graphics~\cite{siren,nerf,sdf,mescheder2019occupancy,idr}. As an alternative to discrete grid-based signal representation, implicit representation is able to parameterize modern signals as samples of a continuous manifold, using multi-layer perceptions (MLP) to map between coordinates and signal values. Several seminal works~\cite{nerf,siren,yariv2020multiview} have verified the effectiveness of INR in representing image, video, and audio. Followups further apply INR to more challenging tasks including novel-view synthesis~\cite{nerf,mipnerf,mipnerf360,zhang2020nerf++}, 3D-aware generative model~\cite{zhou2021cips,chan2021pi,chan2021efficient}, and inverse problem~\cite{chen2021learning,sun2021coil}.
\begin{figure*}
\centering
\begin{tabular}{c c c c c c c}
    \includegraphics[width=0.16\textwidth]{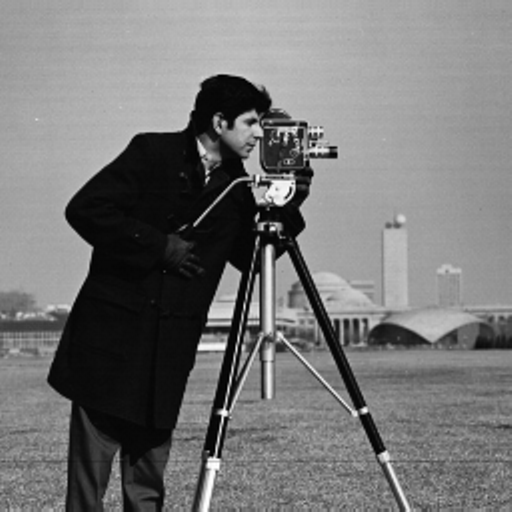} &
   \includegraphics[width=0.16\textwidth]{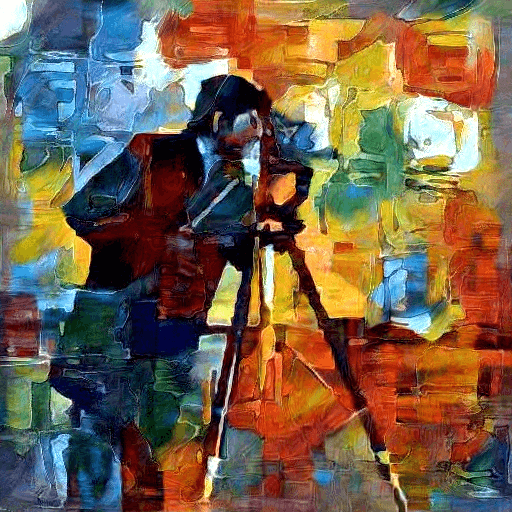} &
   \includegraphics[width=0.16\textwidth]{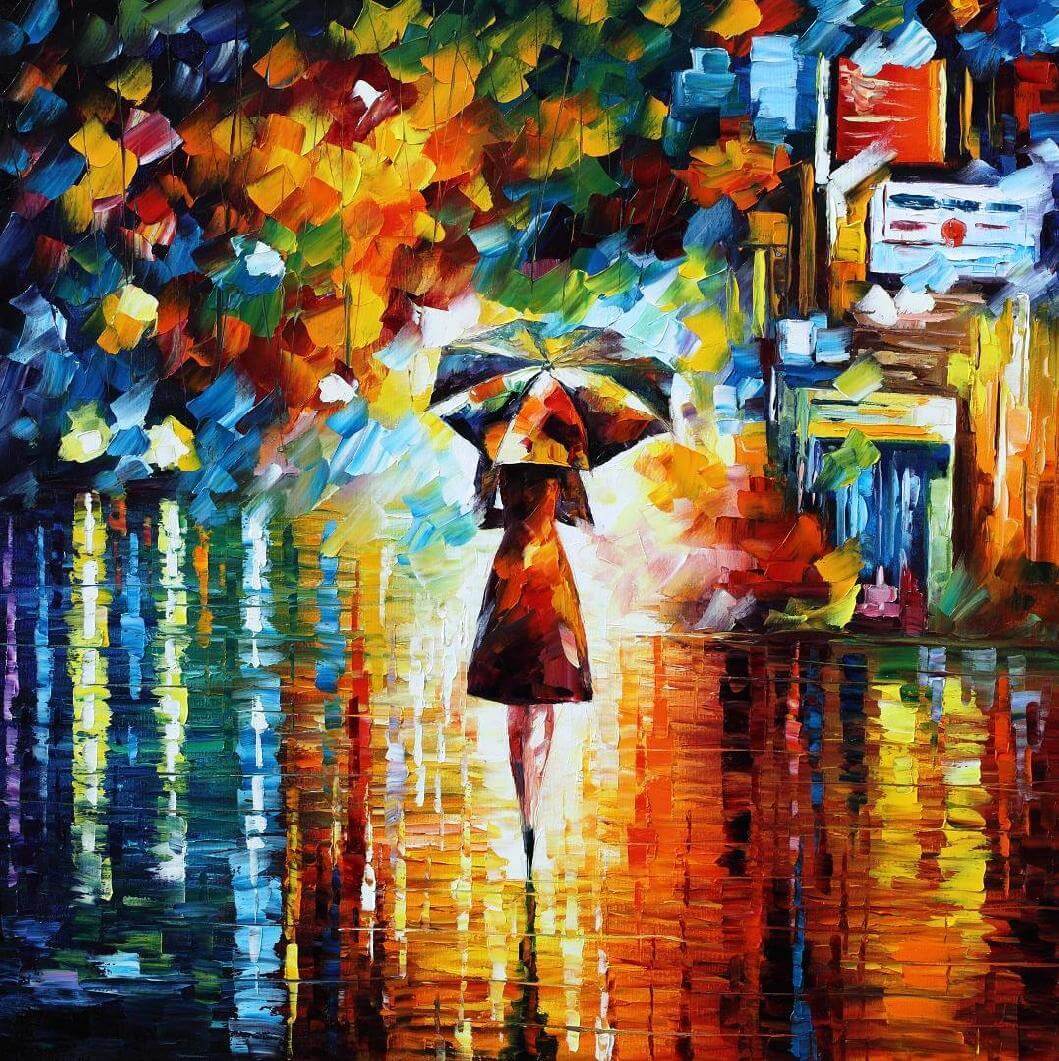} &
   \includegraphics[width=0.16\textwidth]{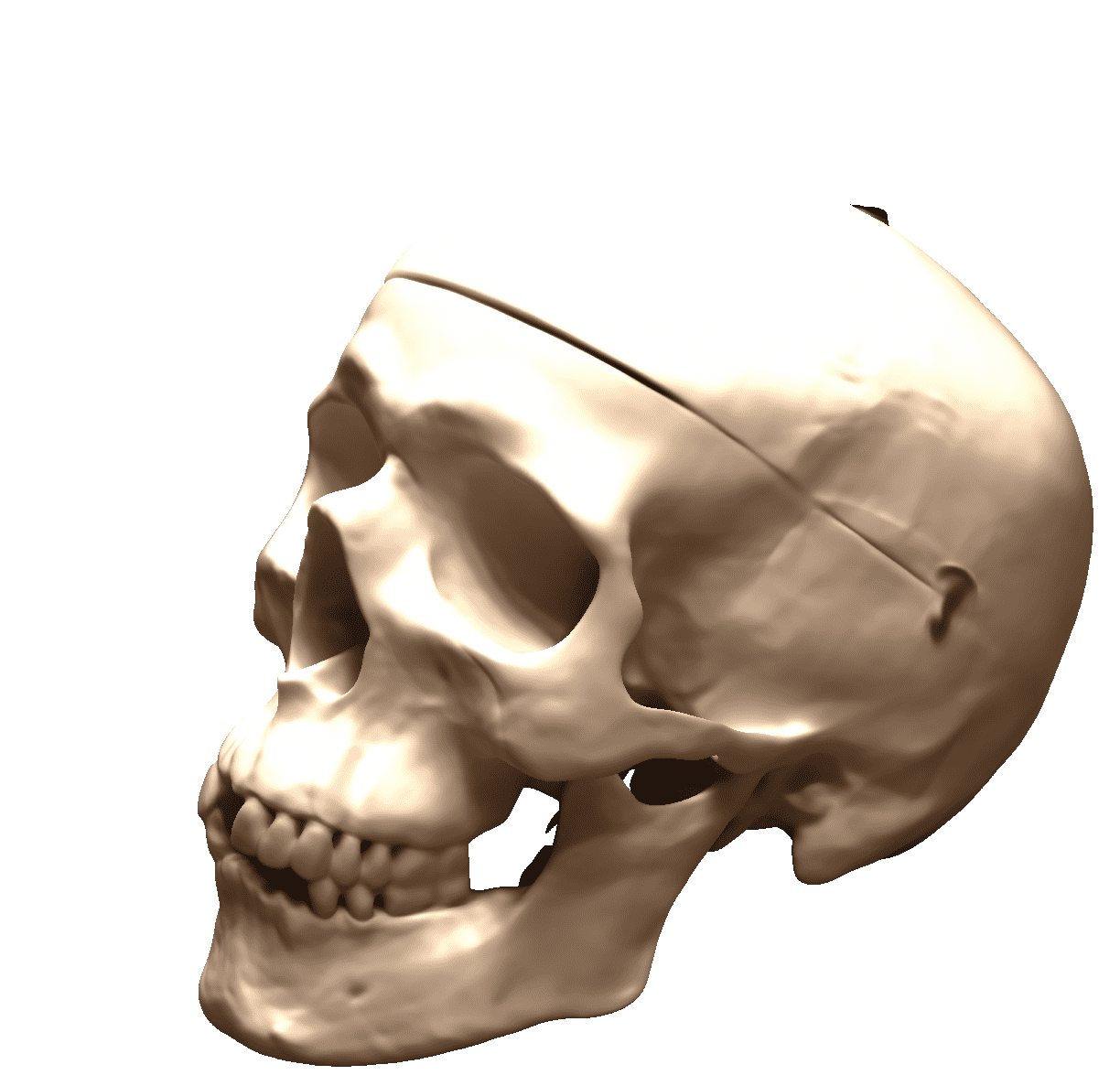} &
   \includegraphics[width=0.16\textwidth]{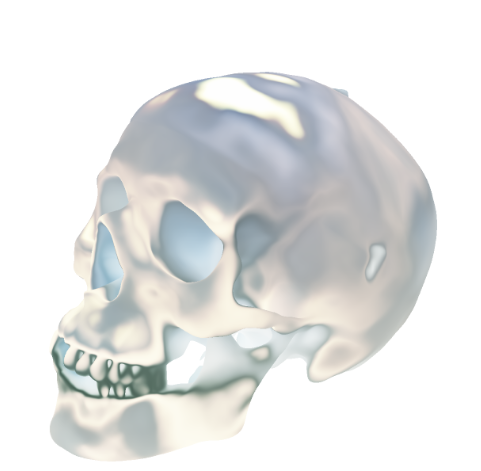}&
   \includegraphics[width=0.16\textwidth]{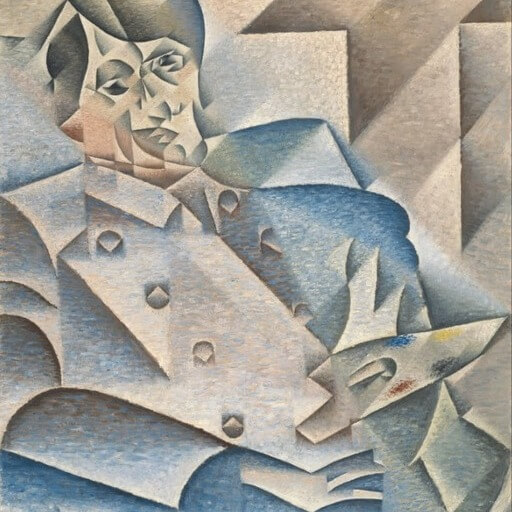}
   \\
   SIREN~\cite{siren} &SIREN+\textbf{INS} & Style Image &SDF~\cite{yariv2020multiview}  &SDF+\textbf{INS} & Style Image \\
    \includegraphics[width=0.16\textwidth]{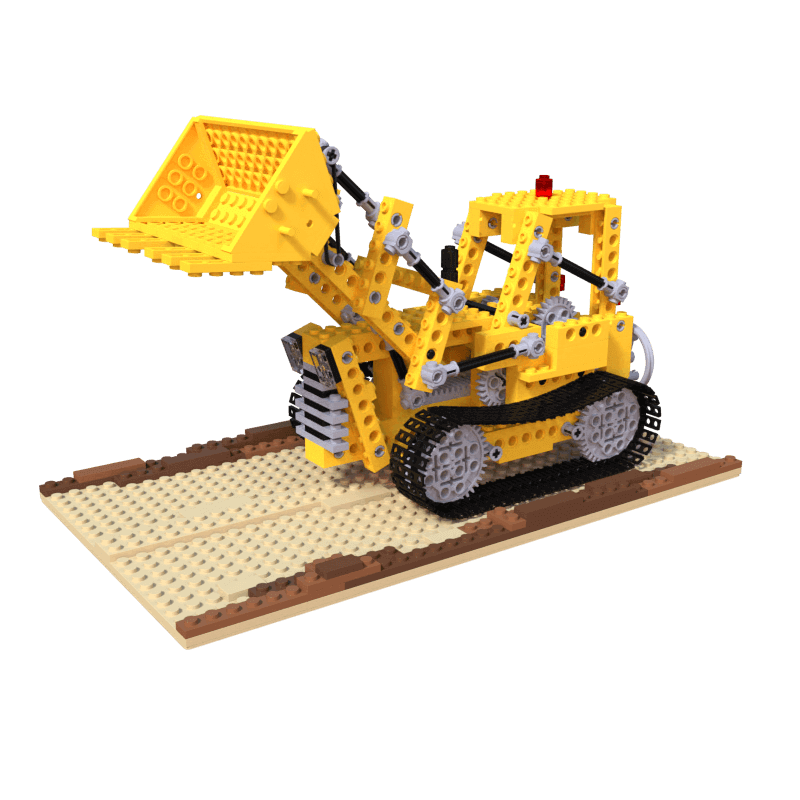} &
    \includegraphics[height=0.16\textwidth]{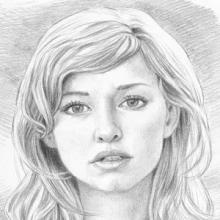} &
   \includegraphics[width=0.16\textwidth]{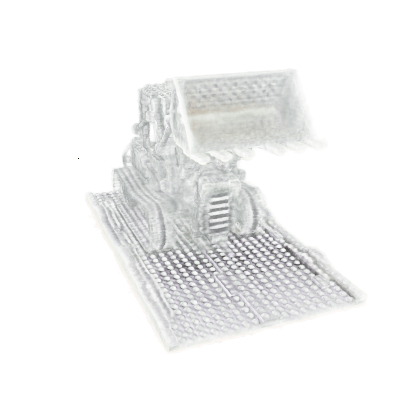} &
   \includegraphics[width=0.16\textwidth]{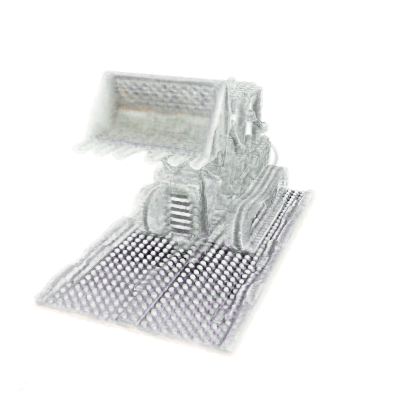} &
  \includegraphics[width=0.16\textwidth]{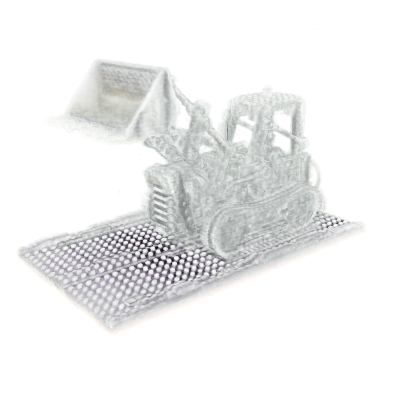} &
  \includegraphics[width=0.16\textwidth]{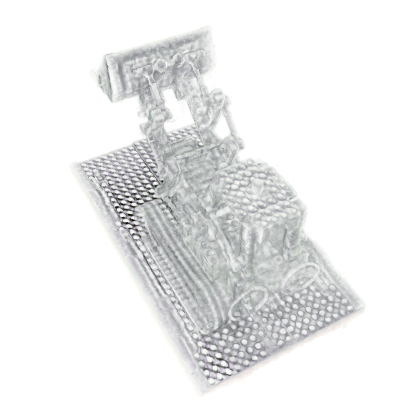} \\
     NeRF~\cite{nerf} &Style Image &NeRF+\textbf{INS} & NeRF+\textbf{INS} &NeRF+\textbf{INS} &NeRF+\textbf{INS}   \\
\end{tabular}
\caption{\textbf{Representative visual examples generated by the proposed method.} We show stylized results on three different types of implicit representation, including 2D coordinate-based mapping function (SIREN~\cite{siren}), Signed Distance Function(SDF~\cite{sdf}), and Neural Radiance Field (NeRF~\cite{nerf}).}\label{fig:teaser}
\end{figure*}
While implicit neural representation reveals multiple advantages compared to conventional discrete signals, a general question of curiosity might be: \textit{which and how modern visual signal processing approaches/tasks designed for discrete signals can also be applied to continuous representations?} 
Research pursuing this answer has been conducted on implicit neural representation since its origin. 
Chen \etal~\cite{liif} apply a local implicit function to image super-resolution and they observe that INR can surpass bilinear and nearest upsampling. Sun \etal~\cite{sun2021coil} demonstrate the effectiveness of INR in the context of sparse-view X-ray CT. Dupont \etal~\cite{dupont2021coin} propose to store the weights of a neural implicit function instead of pixel values, which surprisingly outperforms JPEG compression format.~\cite{zhang2021implicit} further demonstrates superior video compression using similar ideas. 

We investigate a novel setting: to yield visually pleasing stylized examples under various 2D and 3D scenarios, using a generalized approach leveraging implicit neural representations. %
Note that, training a stylized implicit neural representation still faces many hurdles. \underline{On one hand}, the aforementioned works mostly have the access to dense measurements or at least sparse clean data, which enables training an implicit neural network \textbf{under the supervision of target signal}. In contrast to those tasks/approaches, current image stylization mechanisms are mostly conducted in an unsupervised manner, due to the absence of stylized ground truth data. Consequently, it is still unknown whether coordinate-based MLP can be optimized without accessing corresponding ground truth signals. \underline{On the other hand}, existing style images are mostly based on 2D scenes, which raises obstacles when being considered as the appearance of 3D implicit representation. Prior art~\cite{chiang2022stylizing} attempted on marrying stylization with one specific type of Implicit Neural Representation, the neural radiance field (NeRF)~\cite{nerf}.
Nevertheless, it still captures the statistics of style information by a series of pre-trained convolution-based hypernetwork to generate model weights, rather than a direct implicitly encoding stylization. As indicated by recent literature~\cite{lorraine2018stochastic}, training a robust hypernetwork requires a large amount of training samples, while novel-view synthesis tasks commonly hold no more than hundreds of views, potentially jeopardizing the synthesized visual quality.

To conquer the aforementioned fragility, we propose a {U}nified \textbf{I}mplicit \textbf{N}eural \textbf{S}tylization framework, coined as \textbf{INS}.
Different from the vanilla implicit function which is built upon a single MLP network, the proposed framework divides an ordinary implicit neural representation to multiple individual components. Concretely speaking, we introduce a Style Implicit Module to the ordinary implicit representation, and coin the later one as Content Implicit Module in our framework. During the training process, the stylized information and content scene are encoded as one continuous representation, and then fused by another Amalgamation Module. 
To further regularize the geometry of given scenes, we utilize an additional self-distilled geometry consistency loss on top of the rendered density, for the stylization of NeRF. Eventually, INS is able to render view-consistent stylized scenes from novel views, with visually impressive texture details: a few examples are shown in Figure.~\ref{fig:teaser}.

Our contributions are outlined below:
\begin{itemize}
    \item We propose INS, a unified implicit neural stylization framework, consists of a style implicit module, a content implicit module, and an amalgamation module, which enables us to synthesize promising stylized scenes under multiple 2D and 3D implicit representations.
    \item We conduct comprehensive experiments on several popular implicit representation frameworks in this novel stylization setting, including 2D coordinate-based framework (SIREN~\cite{siren}), Neural Radiance Field (NeRF~\cite{nerf}), and Signed Distance Functions (SDF~\cite{sdf}). The rendering results are found to be more consistent, in both shape and style details, from different views. 
    \item We further demonstrate that INS is able to learn representations that are continuous not only with regard to spatial placements (including views), but also in the style space. This leads to effortlessly interpolating between different styles and generating images rendered by the new mixed styles.
    
\end{itemize}

\section{Related Works}

\subsection{Implicit Function}

Recent research has exhibited the potential of Implicit Neural Representation (INR) to replace traditional discrete signals with continuous functions parameterized by multilayer perceptrons (MLP), in computer vision and graphics~\cite{tancik2020fourier,sitzmann2020implicit}. The coordinate-based neural representations~\cite{chen2019learning,mescheder2019occupancy,michalkiewicz2019implicit} have become a popular representation for various tasks such as representing image/video~\cite{siren,dupont2021coin,zhang2021implicit}, 3D reconstruction \cite{genova2019learning,atzmon2019controlling,oechsle2019texture,chen2019learning,gropp2020implicit,mescheder2019occupancy,niemeyer2019occupancy,sdf,peng2020convolutional,saito2019pifu}, and 3D-aware generative modelling \cite{chan2021pi,devries2021unconstrained,gu2021stylenerf,hao2021gancraft,meng2021gnerf,niemeyer2021giraffe,schwarz2020graf,zhou2021cips}.
Analogously, as this representation is differentiable, prior works apply coordinate-based MLPs to many inverse problems in computational photography \cite{chen2021learning,chen2021nerv,sitzmann2021light,attal2021torf,shen2021non,wang2022neural} and scientific computing \cite{li2020fourier,han2018solving,zhong2021cryodrgn}.

\subsection{Implicit 3D Scene Representation}
Traditional 3D reconstruction methods utilizes discrete representations such as point cloud~\cite{aliev2020neural}, meshes~\cite{kato2018neural,shrestha2021meshmvs}, multi-plane images~\cite{mildenhall2019local}, depth maps~\cite{yao2018mvsnet,gu2020cascade} and voxel grids~\cite{kutulakos2000theory}.
Recently, INR has also prevailed among 3D scene representation tasks, which simply adopt an MLP that maps from any continuous input 3D coordinate to the geometry of the scene, including signed distance function (SDF)~\cite{sdf,jiang2020sdfdiff,atzmon2020sal,michalkiewicz2019implicit,gropp2020implicit,sitzmann2019scene,jiang2020local}, 3D occupancy network~\cite{mescheder2019occupancy,chen2019learning}, and so on.
In addition to representing shape, INR has also been extended to encode object appearance.
Among them, Neural Radiance Field (NeRF~\cite{nerf}) is one of the most effective coordinate-based neural representations for photo-realistic view synthesis that represents a scene as a field of particles. Draw inspiration from the preliminary success made by NeRF, a lot of following works further improve and extend it to wider application~\cite{mipnerf,mipnerf360,bergman2021fast,gao2020portrait,martin2021nerf,yu2021pixelnerf,mildenhall2021nerf,park2021hypernerf,chen2022aug,hedman2021baking,mipnerf,mipnerf360,park2021hypernerf,park2021nerfies,xu2022sinnerf}. 
Different from the grid-based approaches, training a stylized implicit representation can not access ground truth signals, which further amplifies the difficulty of optimizing the implicit neural representation.

\subsection{Stylization}
Traditionally, image stylization is formulated as a painterly rendering process through stroke prediction~\cite{zhao2011customizing,zeng2009image}. The first neural style transfer method, proposed by Gatys \textit{et al.} ~\cite{gatys2015neural}, builds an iterative framework to optimize the input image in order to minimize the content and style loss defined by a pre-trained VGG network. Due to the frustratingly large cost of training time, a number of follow-ups further explore how to design a feed-forward deep neural networks~\cite{johnson2016perceptual,ulyanov2016texture}, which obtain real-time performance without sacrificing too much style information. 
Recently, several works extend it to video stylization~\cite{ruder2016artistic,huang2017arbitrary,chen2017coherent,chen2020optical} and 3D environment~\cite{mu20213d,huang2021learning,chiang2022stylizing,Gong_2018_ECCV,hollein2021stylemesh}.

Ha-NeRF~\cite{chen2022hallucinated} is proposed for recovering a realistic NeRF at a different time of day from a group of tourism images, with a CNN to encode the appearance latent code.
The most related NeRF-based stylization works: Style3D~\cite{chiang2022stylizing} and StylizedNeRF~\cite{huang2022stylizednerf} still require a CNN-based hypernetwork or decoder to generate the stylized parameters for neural radiance field. 
In comparison, our proposed INS framework can work for more general implicit representations beyond neural radiance field, and can also be extended to encoding multiple styles. Experiments demonstrate that INS generate more faithful stylization on NeRF compared with Style3D~\cite{chiang2022stylizing}.

\begin{figure*}[t]
\centering
    \includegraphics[width=1\textwidth]{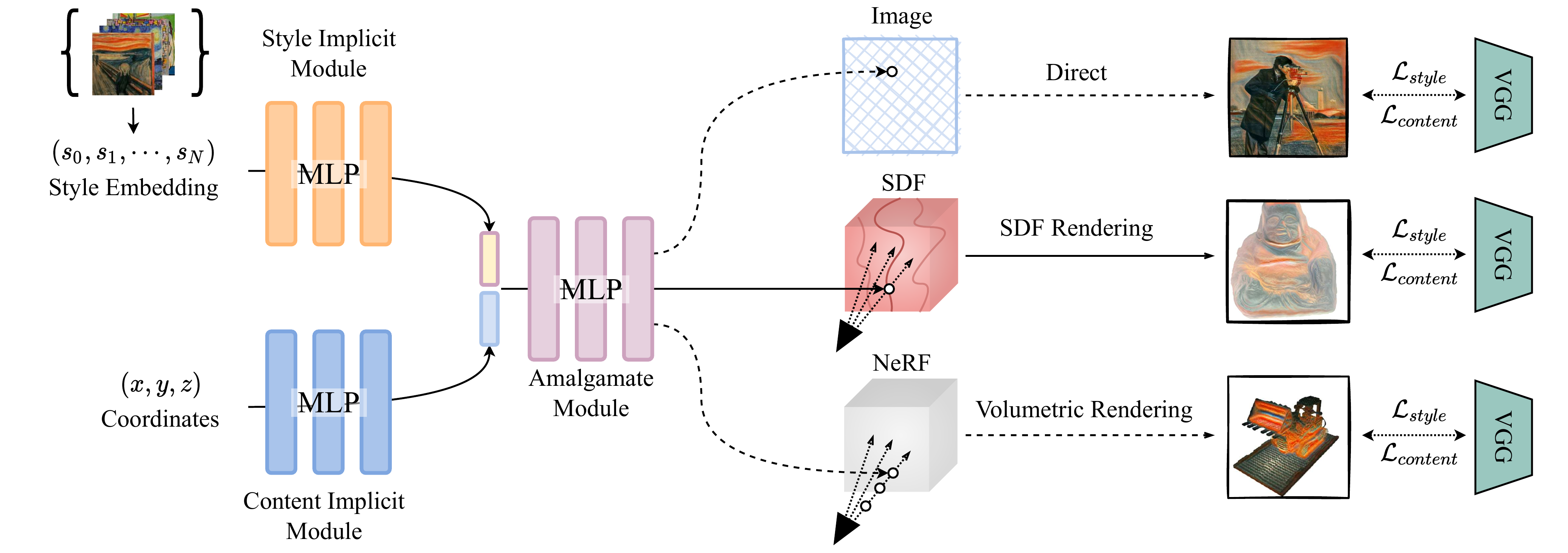}

\caption{\textbf{The main pipeline of unified implicit neural stylization (INS) framework and its components.} We took SDF with the proposed INS, for example, it inputs with implicit coordinates along with ray directions and style embeddings. Style Implicit Module (SIM) and Content Implicit Module (CIM) are used to extract conditional implicit style features and implicit scene features. Amalgamate Module (AM) is applied to fuse features in the two spaces, generating styliezed density and color intensity of each 3D point. An implicit rendering step is applied on the top of AM (i.e., Volume rendering for NeRF, surface rendering for SDF) to render the pixel intensity.
}\label{fig:arc}

\end{figure*}
\section{Preliminary} \label{sec:prelim}
This section introduces the relevant background on several implicit representations and volumetric radiance representations, including image fitting~\cite{siren}, neural radiance field~\cite{nerf} and signed distance function~\cite{sdf}. 
\paragraph{\textbf{Implicit Image Fitting:}} The most prototypical example of neural implicit representation is image regression \cite{siren,tancik2020fourier}.
Consider fitting a function $f: \real^{2} \rightarrow \real^{3}$ that encodes the pixel array of a given image $\Set{I}$ into a continuous representation. Function $f(\Mat{x})$ takes pixel coordinates $\Mat{x} \in \real^2$ as the inputs, and outputs the corresponding RGB colors $\Mat{c} \in \real^3$. Parameterizing $f$ with a multi-layer perception networks (MLPs), it can be optimized by the mean-squared error (MSE) loss function $\mathcal{L} = \E_{\Mat{x} \sim \Prob(\Set{I})} \lVert f(\Mat{x}) - \Mat{c}\rVert_2^2$, where $\Prob(\Set{I})$ is a probability measure support in image lattice $\Set{I}$.

\paragraph{\textbf{Neural Radiance Field:}}
In contrast to point-wisely regression of implicit fields, NeRF~\cite{nerf} proposes to reconstruct a radiance field by inversing a differentiable rendering equation from captured images.
Specifically, NeRF learns an MLP $f: (\Mat{x}, \Mat{\theta}) \mapsto (\Mat{c}, \sigma)$ with parameters $\Mat{\Theta}$, where $\Mat{x}$ is the spatial coordinate in 3D space and $\Mat{\theta}$ represents the view directions $\in [-\pi, \pi]^2$. The output $\Mat{c} \in \real^{3}$ indicates the predicted color of the sampled point, $\sigma \in \real_{+}$ signifies its density value.
The pixel color intensity can be obtained using volume rendering~\cite{drebin1988volume} by ray tracing, integrating the predicted color and density along the ray.
To render a pixel on the image plane, NeRF casts a ray $\Mat{r} = (\Mat{o}, \Mat{d}, \Mat{\theta})$ through the pixel and accumulate the color and density of $K$ point samples along the view direction in the 3D space. The pixel color intensity can be estimated:
\begin{align} \label{eqn:vol_render}
& \Mat{C}(\Mat{r} \vert \Mat{\Theta}) = \sum_{k=1}^{K} T_k (1 - \exp(-\sigma_{k} \Delta t_k)) \Mat{c}_{k},
\end{align}
where $(\Mat{c}_k, \sigma_k) = f(\Mat{x}_k, \Mat{\theta})$, $\Mat{x}_k = \Mat{o} + t_k \Mat{d}$, $t_k$ are the marching distance of sampled points, and $T_k = \exp( -\sum_{l=1}^{k-1} \sigma_{l} \Delta t_l)$ is known as the transmittance to model occlusion.
$\Delta t_k = t_{k+1} - t_k$ indicates the distance of sampled point in 3D space.
With this approximated rendering pipeline, the model weights are optimized by minimizing the $L_2$ distance between rendered ray colors $\Mat{C}(\Mat{r})$ and captured pixel colors $\widehat{\Mat{C}}(\Mat{r})$ as follows:
\begin{align} \label{eqn:photo_loss}
\Mat{\Theta}^* = \argmin_{\Mat{\Theta}} \E_{\Mat{r} \sim \Prob(\Set{R})} \left\lVert \Mat{C}(\Mat{r} \vert \Mat{\Theta}) - \widehat{\Mat{C}}(\Mat{r}) \right\rVert_2^2,
\end{align}
where $\Set{R}$ is a collection of rays cast from all pixels in the training set, and $\Prob(\Set{R})$ defines a distribution over it.

\paragraph{\textbf{Implicit Surface Representation:}}
Signed Distance Function (SDF)~\cite{curless1996volumetric} $f: \real^3 \rightarrow \real$ is an implicit representation of 3D geometries. SDF specifies each spatial point with the signed distance to the implicit iso-surface, where the sign indicates whether the point is inside or outside the object.
Recent works of \cite{park2019deepsdf,sitzmann2020metasdf} propose to employ MLPs to represent this continuous field via direct supervision using point clouds.
To optimize a textured SDF from multi-view images like NeRF \cite{mildenhall2021nerf}, Yariv \etal \cite{yariv2020multiview} proposes a neural rendering pipeline, named IDR, which enables rendering images from an SDF.
With this framework, one can indirectly supervise SDF using its multi-view projections.
Suppose given a camera pose, we can cast rays $\Mat{r}=(\boldsymbol{o}, \boldsymbol{v})$ through each pixel to trace an intersected point with the surface:
\begin{align}
    \Mat{\hat{x}} = \Mat{o} + t_0 \Mat{v} - \frac{\Mat{v}}{\nabla f_0 \cdot \Mat{v}_0} f(\Mat{o} + t_0 \Mat{v}),
\end{align}
where $t_0$, $\Mat{v}_0$ and $f_0$ are initial states when performing ray tracing (see \cite{yariv2020multiview}). After obtaining the intersection $\Mat{\hat{x}}$ of ray and surface, IDR also lets the SDF network $f$ output an appearance embedding $\Mat{\hat{\gamma}}$, and computes the normal $\Mat{\hat{n}} = \nabla f(\Mat{\hat{x}})$. Then the ray color can be rendered by another rendering MLP conditioned on both point coordinate $\Mat{\hat{x}}$ and normal $\Mat{\hat{n}}$:
\begin{align}
    \Mat{C}(\Mat{r} \lvert \Mat{\Theta}) = r(\Mat{\hat{x}}, \Mat{\hat{n}}, \Mat{d}, \Mat{\hat{\gamma}}).
\end{align}
Similar to NeRF \cite{mildenhall2021nerf}, $f$ and $r$ are simultaneously optimized by photometric loss between captured image pixels and rendered rays (see Equation~\ref{eqn:photo_loss}).

\section{Method}
We next illustrate the main pipeline of Unified \textbf{I}mplicit \textbf{N}eural \textbf{S}tylization (INS). INS consists of a Style Implicit Module (SIM) to transform the input style embedding into implicit style representations, a Content Implicit Module (CIM) to map the input coordinates into implicit scene representations, and an Amalgamate Module (AM) which amalgamates the two representation to predict RGB intensity. 
To preserve the geometry fidelity while generating the stylized texture of rendered views, a self-distilled geometry consistency regularization is applied upon the INS framework.

\subsection{Implicit Style and Content Representation}
Generating the stylized images $\Mat{Y}$ can be formulated as an energy minimization problem~\cite{gatys2015neural}. It consists of a \textit{content loss} and a \textit{style loss}, defined under a pre-trained VGG network~\cite{simonyan2014very}. We build upon the prior work and thus propose our implicit stylization framework for SIREN, SDF and NeRF.
\paragraph{\textbf{Content Representation}}
The content loss in 2D images stylization pipeline~\cite{gatys2015neural} $\mathcal{L}_{\mathrm{content}}$ is defined as:
\begin{align}
\mathcal{L}_{\mathrm{content}}(\Mat{C}, \Mat{Y}) = \frac{1}{C_{i,j}W_{i,j}H_{i,j}}  \sum_{(i,j) \in \Set{J}} \left\lVert F_{i,j}(\Mat{C}) - F_{i,j}(\Mat{Y}) \right\rVert_F^2,
\end{align}
where $\Mat{C}$ denotes the content ground truth image and $\Mat{Y}$ denotes synthesized output, $F_{i,j}$ denotes the feature map extracted from a VGG-16 model pre-trained on ImageNet, $i$ represents its $i$-th max pooling, and $j$ represents its $j$-th convolutional layer after $i$-th max pooling layer. $C_{i,j}$, $W_{i,j}$ and $H_{i,j}$ are the dimensions of the extracted feature maps. We adapt the content loss to the intermediate layer of INS pipeline to preserve the content of the predicted color image patch, we choose $i$ = 2, $j$ = 2 by default.

\paragraph{\textbf{Style Representation}}
To extract representation of the stylized information, ~\cite{simonyan2014very} introduces a different feature space to capture texture information~\cite{gatys2015texture}. Similar to the content loss, the feature space is built upon the filter response in multiple layers of a pre-trained VGG network. By capturing the correlations of the filter responses expressed by the Gram matrix $\Mat{G}_{i,j} \in  \real^{C_{i,j} \times C_{i,j}}$ between the style image $\Mat{S}$ and the synthesized image $\Mat{Y}$, multi-scale representations can be obtained to capture the texture information from the style image and endow such texture on the stylized image. Here, we define our style loss $\mathcal{L}_{\mathrm{style}}$ using the same layers of VGG-16 with~\cite{gatys2015neural} on the top of the prediction of implicit neural representations and the given style image:
\begin{align}
& \mathcal{L}_{\mathrm{style}}(\Mat{S}, \Mat{Y}) = \sum_{(i,j) \in \Set{J}} \left\lVert \Mat{G}_{i,j}(\Mat{S}) - \Mat{G}_{i,j}(\Mat{Y}) \right\rVert_F^2,\\
& \text{where } [\Mat{G}_{i,j}]_{c,c'}(\Mat{Y}) = \frac{1}{C_{i,j}W_{i,j}H_{i,j}} \sum_{k=1}^{H \times W} F_{i,j}(\Mat{Y})_{c,k} F_{i,j}(\Mat{Y})_{c',k},
\end{align}
where $\Set{J}$ are the indices of selected feature maps.
In practice, we choose $\Set{J} = \{(1, 2), (2, 2), (3, 3), (4, 3)\}$ in our experiments.

\paragraph{\textbf{Conditional INS Stylization}}
Conditional encoding has been widely applied in convolutional networks~\cite{iizuka2016let,yanai2017conditional}. Similarly, we propose the conditional implicit representation by input with style conditioned embeddings using and extracting style-dependent features to render stylized color and density, which is shown in Figure~\ref{fig:arc}.
In training, we prepare $n$ style images with an $n$ dimensional one-hot style-condition vector. A mini-batch is constructed with the combinations of one content training patch and all candidate style images. The one-hot vector is fed into SIM to extract the $w$-dimensional style features, they are then concatenated with the implicit representations output from CIM. The following layers of AM take the two features, aggregate them to render the pixel intensity and scene geometry along the rays. A pre-trained VGG~\cite{simonyan2014very} is appended on the top of the INS pipeline to apply implicit style and content constraints during training. During the inference stage, we discard the VGG network, the INS framework becomes a pure MLPs-based network.

\begin{wrapfigure}{r}{0.50\textwidth}
 \centering
\includegraphics[width=0.50\textwidth]{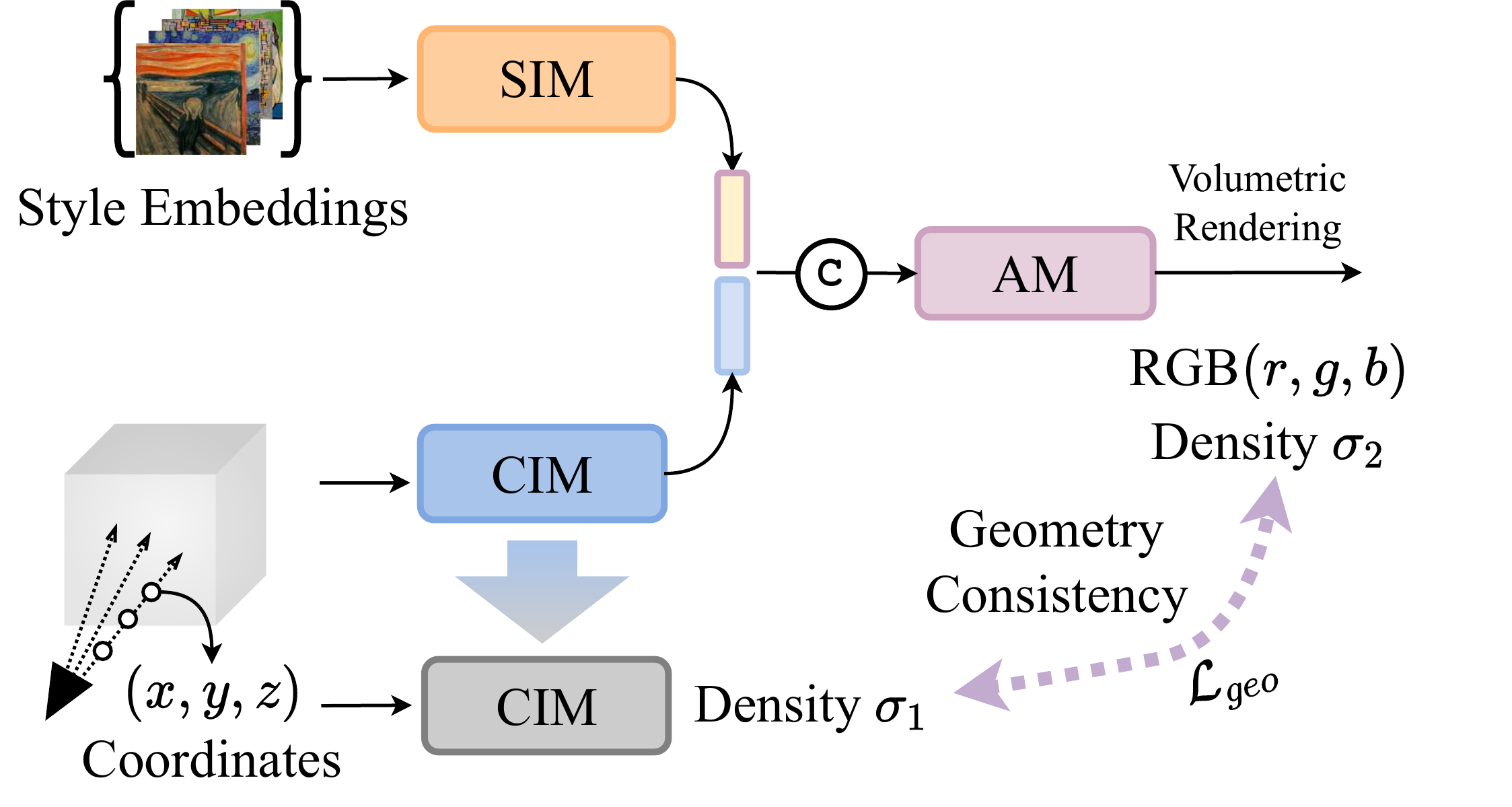}
\par
\caption{\textbf{The proposed self-distilled geometry consistency.} The CIM weights in the grey box are from a pre-trained NeRF and are kept fixed since then. During fine-tuning, the output density of the fixed CIM serves as geometry constrains for the stylized density from the output of AM.}
\label{fig:arc_nerf}
\end{wrapfigure}

\subsection{Geometry Consistency for Neural Radiance Field}
NeRF~\cite{nerf} learns reasonable 3D geometry inherently due to the particular design of implicit radiance field and the supervision from multiple views. However, the INS framework is expected to integrate style statistics from 2D image into 3D radiance field, where no multi-view style images accessible during training process.
To specialize INS for neural radiance fields, we propose to regularize INS with proper geometry constraint to produce faithful shape and appearance.
As the ground truth of target geometry is unavailable in most novel view synthesis benchmarks, we seek help from the self-distillation framework~\cite{xu2020improving}. Concretely speaking, we first train the content implicit module (CIM) only to obtain a clean geometry $\sigma_1$, following the vanilla NeRF training pipeline. After that, we copy the trained weight of CIM and keep that fixed (as shown in the grey block in Figure~\ref{fig:arc_nerf}). In the next, we turn to optimize the whole INS framework (SIM, CIM and AM) with implicit stylization constrains. Meanwhile, a self-distilled geometry constraint between the original geometry $\sigma_1$ produced by the fixed CIM weight and the final stylized geometry $\sigma_2$ reconstructed by the implicit neural stylization framework.
The objective of self-distilled geometry consistency loss is formulated as $\mathcal{L}_{\mathrm{geo}} = \lvert \sigma_1 - \sigma_2 \rvert$, where we adopt the mean-absolute error for the densities of each sampled point.

\paragraph{\textbf{Sampling-Stride Ray Sampling}}
\label{sec_Sampling stride Ray Sampling}
Neural Radiance Field casts a number of rays (typically not adjacent) from camera origin, intersecting the pixel, into the volume and accumulating the color based on density along the ray. 

While our model input with rays intersected with an image patch of size $\mathcal{P} \in K \times K$, predicting the stylized patch $\mathcal{P'}\in K \times K$ with its texture closed to the given style images. 2D style transfer methods~\cite{chen2017coherent,huang2017arbitrary} typically crop the patch larger than $256\times 256$. However, it is too expensive for the neural radiance field as it queries the MLPs more than $256 \times 256 \times N$ times of the MLP for each step~\cite{nerf}, where $N$ indicates sampled points number along each ray. 

Similar to~\cite{schwarz2020graf,meng2021gnerf}, we adopt a $\textit{Sampling-Stride Ray Sampling}$ strategy to enlarge the receptive field of the sampled patch to capture a more global context. The illustration of the ray sampling can be found in our supplementary materials, where a sampling stride larger than 1 result in a large receptive field while keeping computational cost fixed.
\subsection{Optimization}
Let $\Mat{F}(\Mat{T})$ denote an INS model parameterized by $\Mat{\Theta}$, which synthesizes an image (patch) from the view specified by the camera pose $\Mat{T} \in \real^{4 \times 4}$ by marching and rendering the rays for all the pixels on the image (patch).
Given multi-view images and the corresponding camera parameters $\Set{T} = \{\Mat{C}_i, \Mat{T}_i\}_{i=1}^{N}$, as well as a set of style images $\Set{S} = \{ \Mat{S}_i \}_{i=1}^{M}$, we train the INS using a combination of losses including reconstruction loss $\mathcal{L}_\mathrm{recon}$, geometry consistency loss $\mathcal{L}_\mathrm{geo}$,
content loss $\mathcal{L}_\mathrm{content}$ and style loss $\mathcal{L}_\mathrm{style}$:
\begin{align}
& \mathcal{L}_\mathrm{total}(\Mat{\Theta} \vert \Set{T}, \Set{S}, \Mat{\Theta}_{\mathrm{vgg}}) = \E_{(\Mat{C}, \Mat{T}) \sim \Prob(\Set{T})} \left[\lambda_0 \mathcal{L}_\mathrm{recon}(\Mat{F}(\Mat{T}), \Mat{C})
+ \lambda_1 \mathcal{L}_\mathrm{geo}(\Mat{F}(\Mat{T}))\right] \\
& + \E_{\Mat{S} \sim \Prob(\Set{S})} \E_{(\Mat{C}, \Mat{T})\sim \Prob(\Set{T})} \left[ \lambda_2 \mathcal{L}_\mathrm{content}(\Mat{F}(\Mat{T}), \Mat{C}|\Mat{\Theta}_{\mathrm{vgg}}) + \lambda_3 \mathcal{L}_\mathrm{style}(\Mat{F}(\Mat{T}), \Mat{S} \vert \Mat{\Theta}_{\mathrm{vgg}}) \right], \nonumber
\end{align}
where $\lambda_0$, $\lambda_1$, $\lambda_2$, $\lambda_3$ control the strength of each loss term, $\Mat{\Theta}_{\mathrm{vgg}}$ denotes the parameters of the VGG network, and $\Prob(\cdot)$ defines a distribution over a support.

\section{Experiments}

\subsection{Stylization on SIREN MLPs} 

\begin{figure*}
\centering
\begin{tabular}{c c c c c c}
    \includegraphics[width=0.19\textwidth]{figures/siren_cam_gt.png} &
   \includegraphics[width=0.19\textwidth]{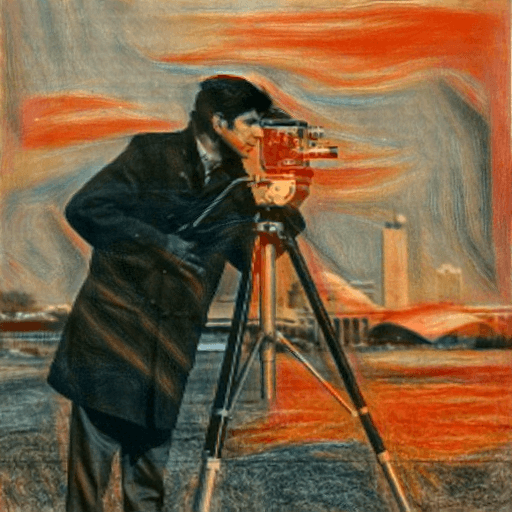} &
   \includegraphics[height=0.19\textwidth]{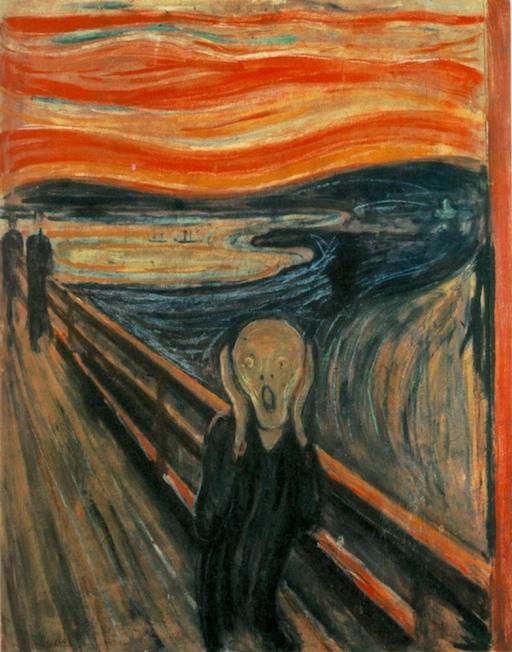} &
   \includegraphics[width=0.19\textwidth]{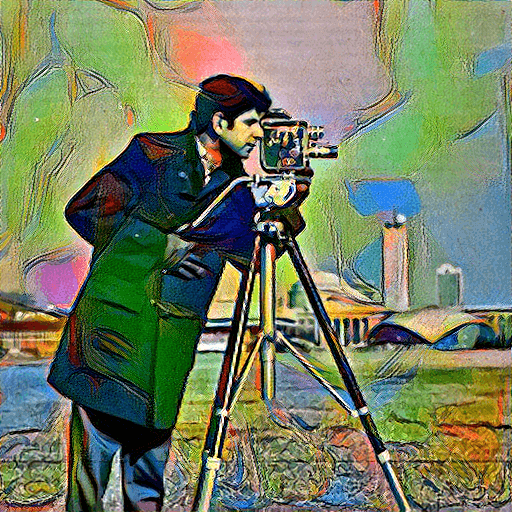} &
   \includegraphics[height=0.19\textwidth]{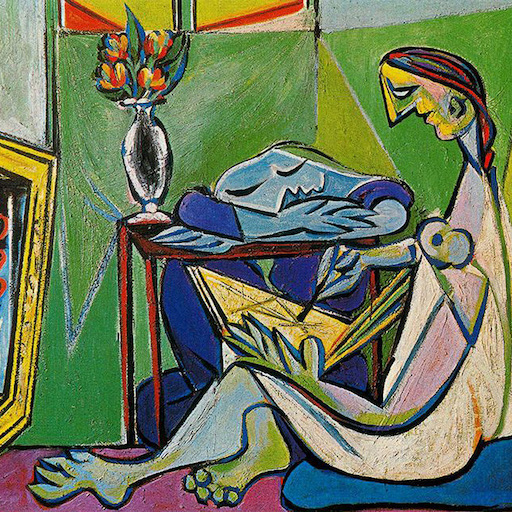} \\
   \includegraphics[width=0.19\textwidth]{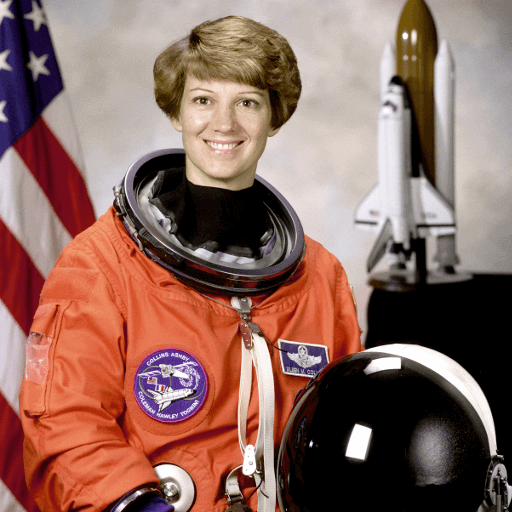} &
   \includegraphics[width=0.19\textwidth]{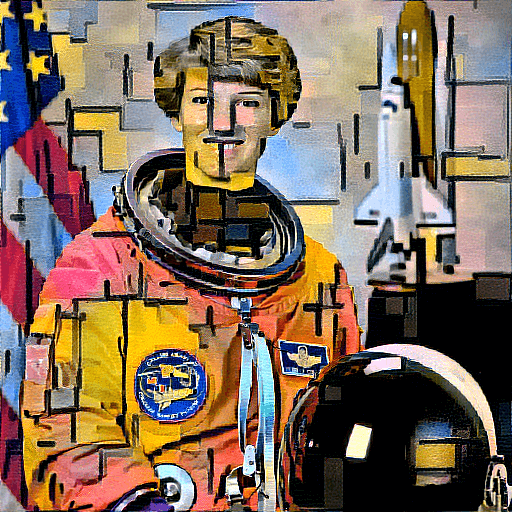} &
   \includegraphics[height=0.19\textwidth]{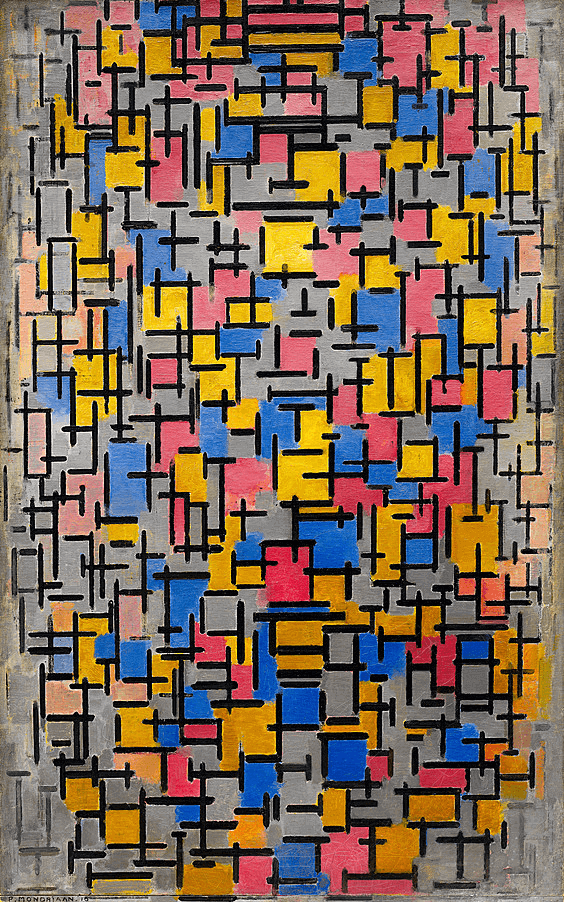} &
   \includegraphics[width=0.19\textwidth]{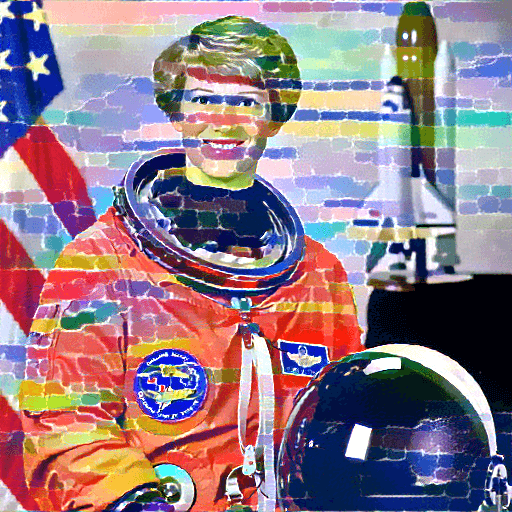} &
   \includegraphics[height=0.19\textwidth]{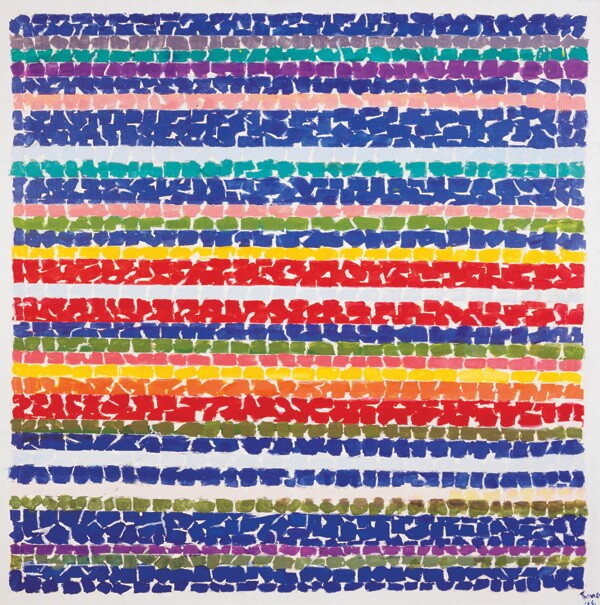} \\
     SIREN &SIREN+INS &Style Image & SIREN+INS &Style Image \\
\end{tabular}
\caption{\textbf{Visual examples generated by applying INS to SIREN~\cite{siren}.} Given an input image used for fitting and style images, SIREN+INS can express the style statistics via an implicit manner(i.e., MLPs).}\label{fig:siren}
\end{figure*}

As one representative example, we apply INS on fitting an image via SIREN~\cite{sitzmann2020implicit} MLPs. %
We reuse the original SIREN framework~\cite{siren} as CIM and follow its training recipes to fit images of 512$\times$512 pixels. Besides that, we also incorporate the SIM and AM on SIREN. A pre-trained VGG-16 network is appended on the output to provide style and content supervisions during training. As seen in Figure~\ref{fig:siren}, the proposed framework successfully in representing the images with the given style statistics in an implicit way.

\begin{figure*}
\centering
\begin{tabular}{c c c c c }

   \includegraphics[width=0.19\textwidth]{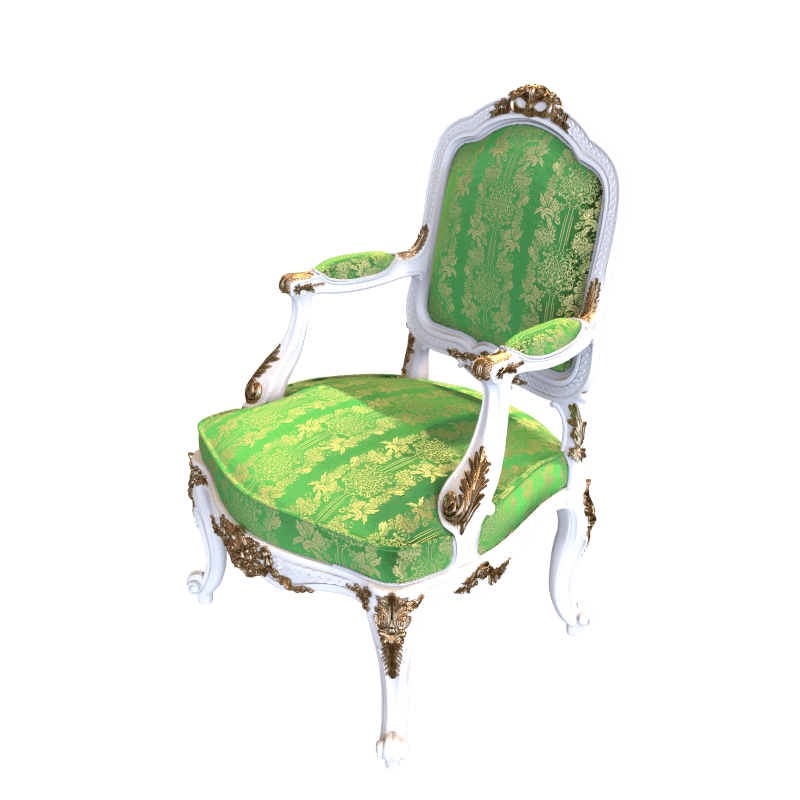} &
   \includegraphics[height=0.19\textwidth]{figures/the_scream.jpg} & 
   \includegraphics[width=0.19\textwidth]{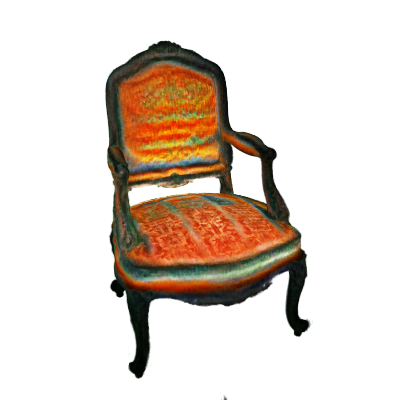} &
   \includegraphics[width=0.19\textwidth]{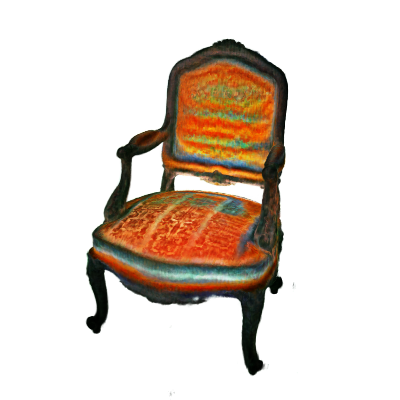} &
   \includegraphics[width=0.19\textwidth]{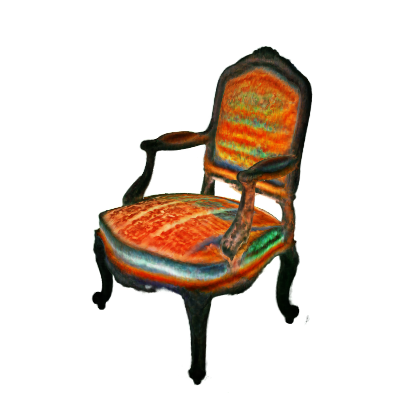}\\
   Color Image &Style Image & INS(view1) &INS(view2) &INS(view3)   \\
   \includegraphics[width=0.19\textwidth]{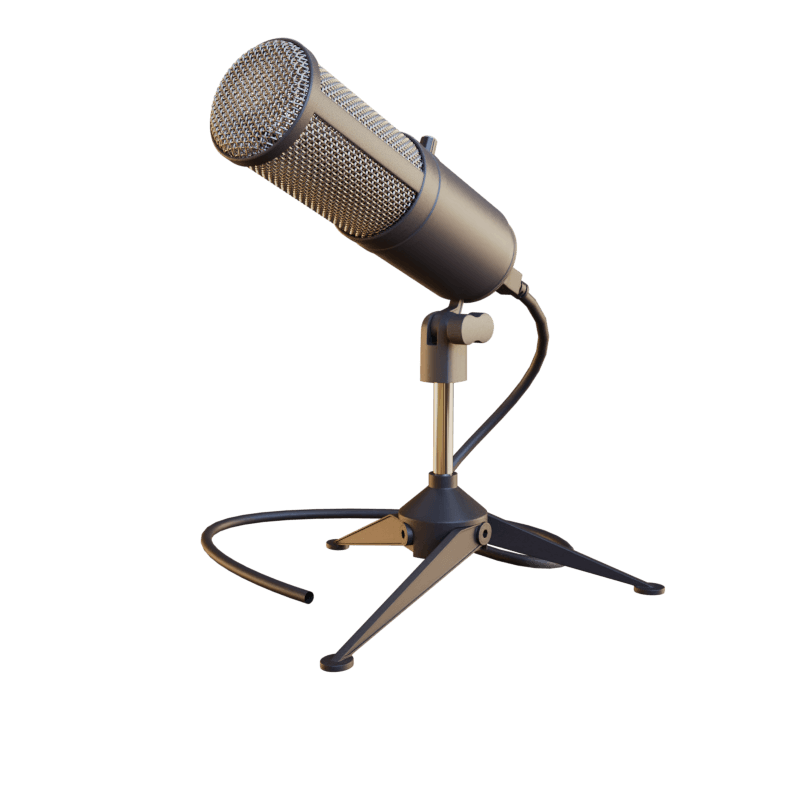} &
   \includegraphics[width=0.19\textwidth]{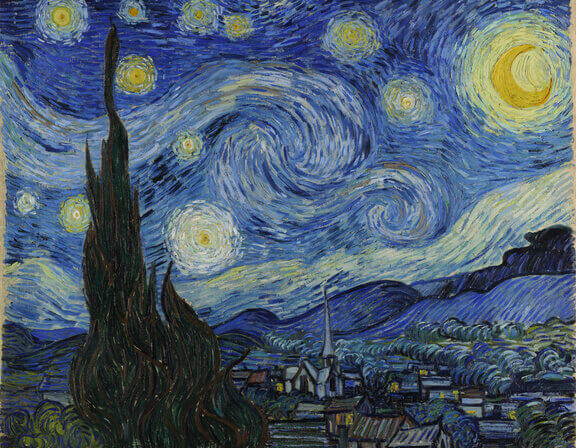} & 
   \includegraphics[width=0.19\textwidth]{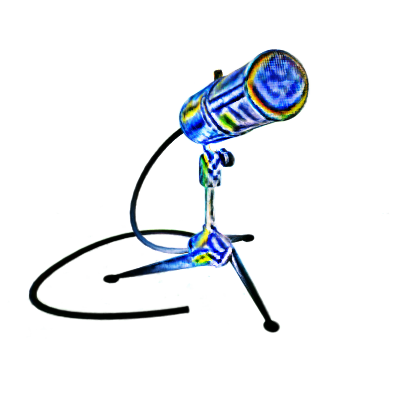} &
   \includegraphics[width=0.19\textwidth]{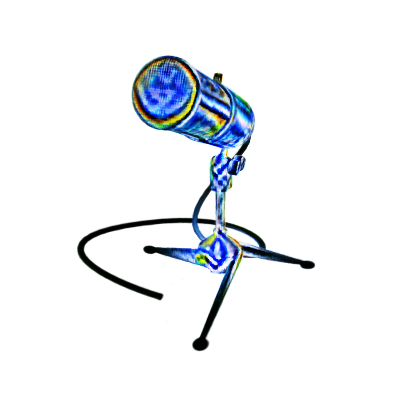} &
   \includegraphics[width=0.19\textwidth]{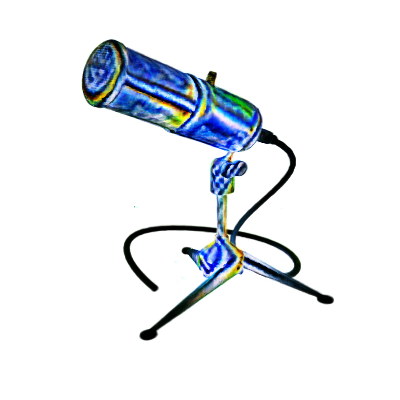}\\
   Color Image &Style Image & INS(view1) &INS(view2) &INS(view3)   \\
   \includegraphics[width=0.19\textwidth]{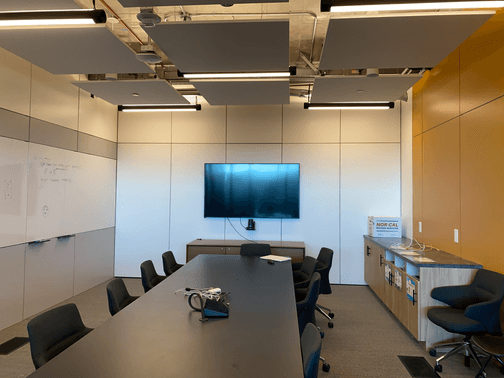} &
   \includegraphics[height=0.15\textwidth]{figures/style_1.png} & 
   \includegraphics[width=0.19\textwidth]{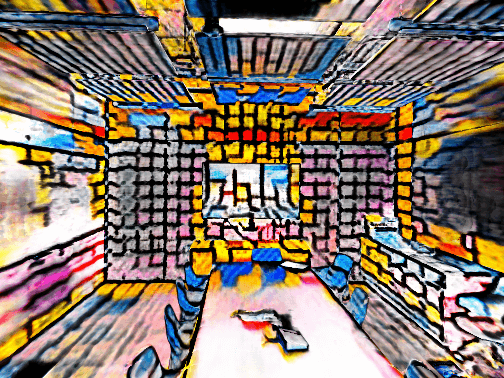} &
   \includegraphics[width=0.19\textwidth]{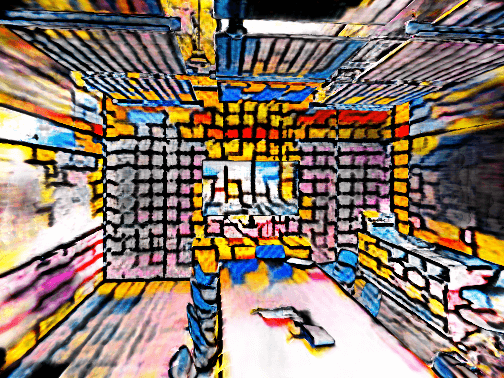} &
   \includegraphics[width=0.19\textwidth]{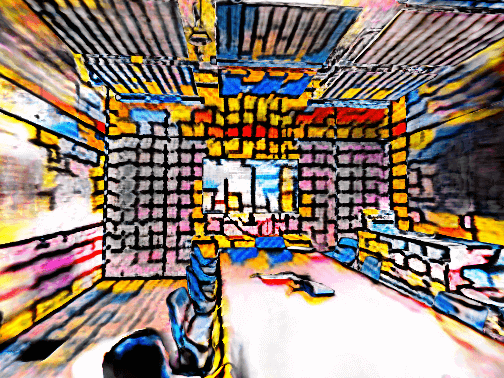}\\
   Color Image &Style Image & INS(view1) &INS(view2) &INS(view3)   \\
   \end{tabular}
\caption{\textbf{Multi-view examples rendered by applying INS to the neural radiance field.} The rendered scenes and objects show consistent stylized texture under different views.}\label{fig:sole_exp_on_nerf}
\end{figure*}
\subsection{Novel View Synthesis with NeRF}

\begin{figure}
\centering
\begin{tabular}{c c c c c}
   \includegraphics[width=0.19\textwidth]{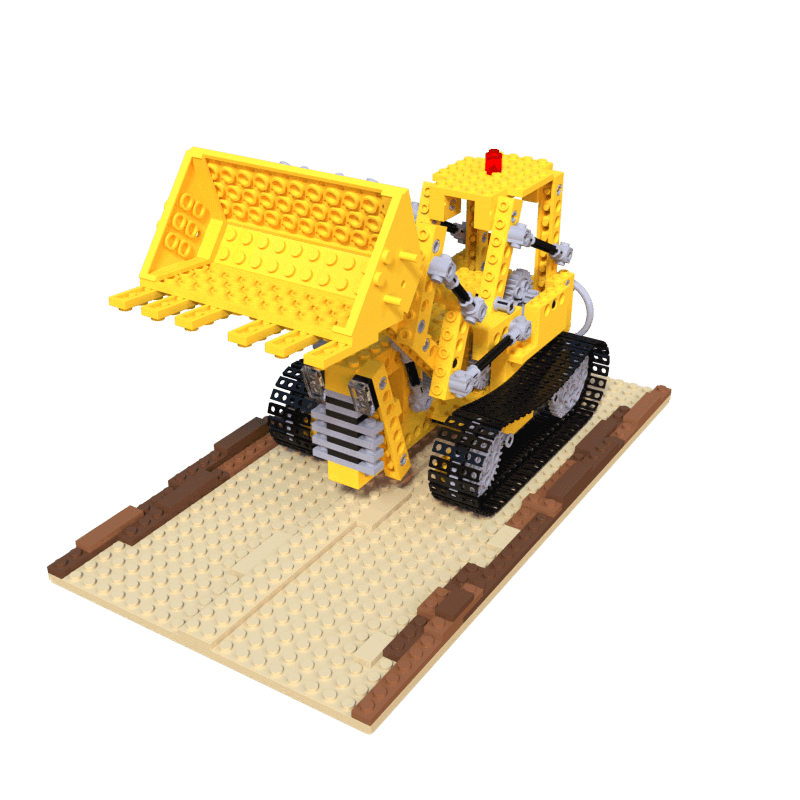} &
      \includegraphics[width=0.19\textwidth]{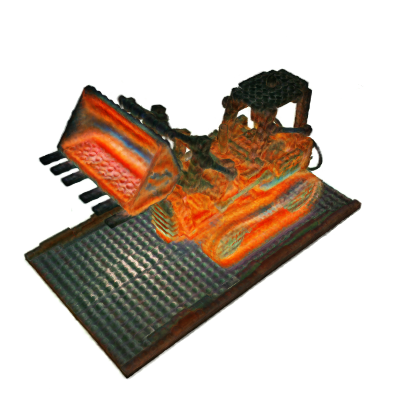} &
  \includegraphics[width=0.19\textwidth]{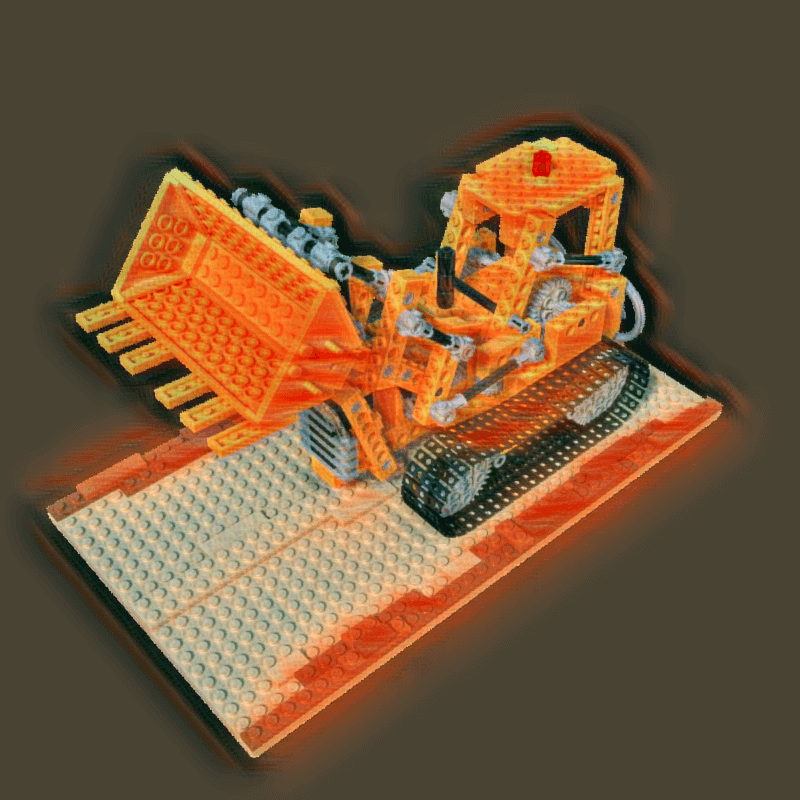} & 
   \includegraphics[width=0.19\textwidth]{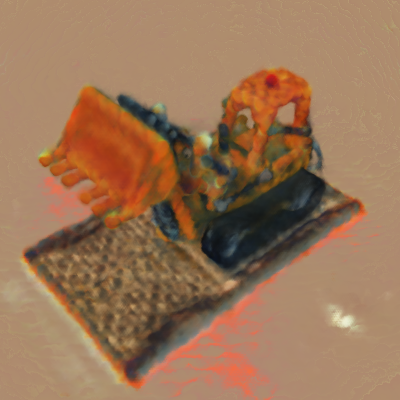} & 
   \includegraphics[width=0.19\textwidth]{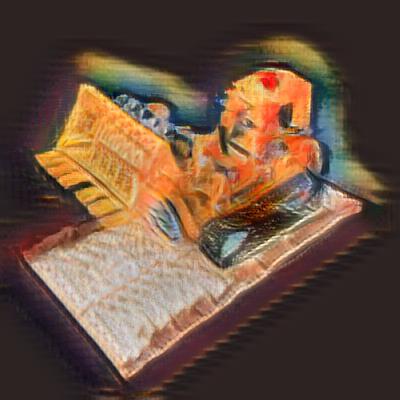} 
\\
   \scriptsize{Color Image}  
   &\scriptsize{Ours(view1)}  
   &\scriptsize{~\cite{johnson2016perceptual}(view1)}  
    &\scriptsize{Style3D~\cite{chiang2022stylizing}(view1)} 
    &\scriptsize{Adain~\cite{huang2017arbitrary}(view1)}\\
    \includegraphics[height=0.18\textwidth]{figures/the_scream.jpg} &
    \includegraphics[width=0.19\textwidth]{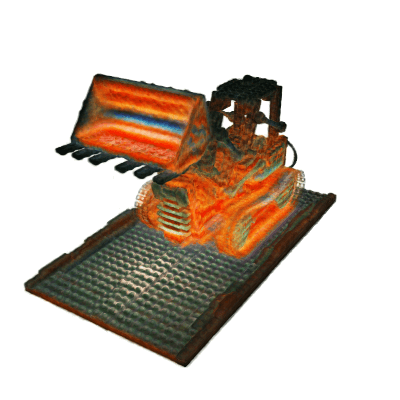} &
   \includegraphics[width=0.19\textwidth]{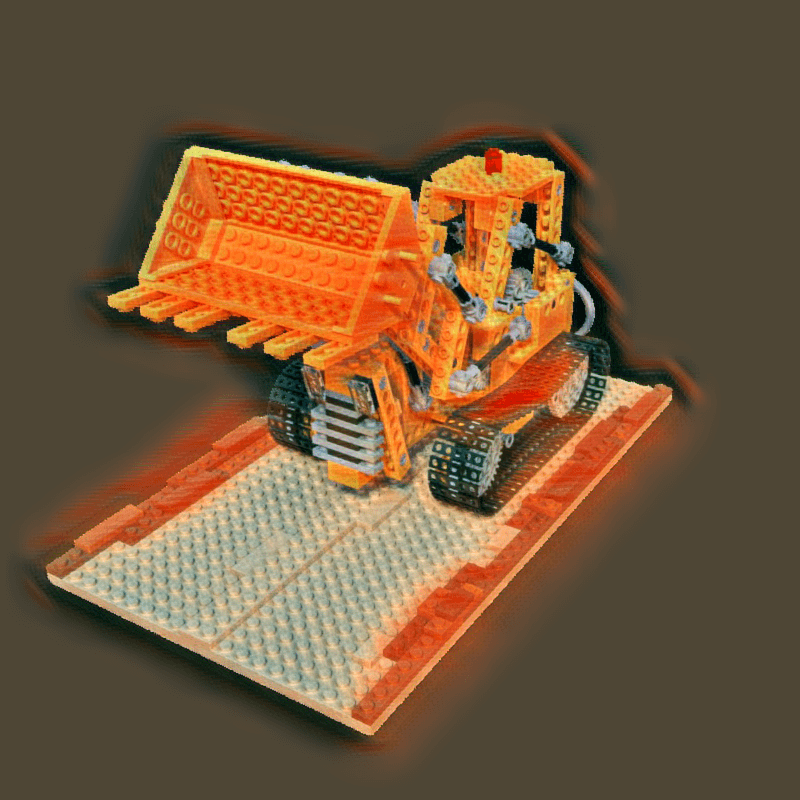}  & 
   \includegraphics[width=0.19\textwidth]{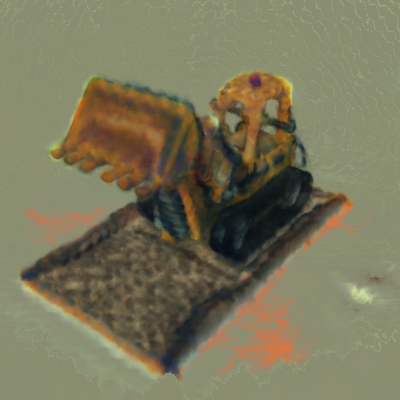} &
   \includegraphics[width=0.19\textwidth]{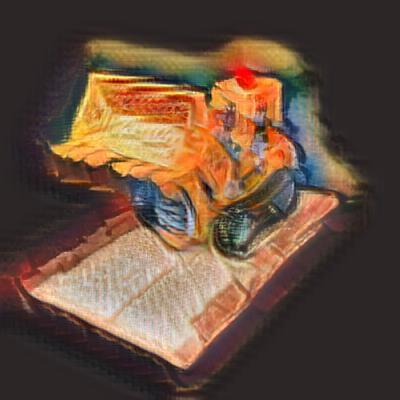}  
\\
   \scriptsize{Style Image}  
   &\scriptsize{Ours(view2)}
   &\scriptsize{~\cite{johnson2016perceptual}(view2)}
   &\scriptsize{Style3D~\cite{chiang2022stylizing}(view2)}
   &\scriptsize{Adain~\cite{huang2017arbitrary}(view2)} 
   \\
\end{tabular}
\caption{Qualitative comparisons. We compare INS with several methods on novel view synthesis datasets. 
Three scenes with different style images are demonstrated, we can see our proposed INS method preserves the learned geometry better (e.g. with clean background) and capture global context while achieving the desired styles.
}\label{fig:compare}
\end{figure}
\paragraph{\textbf{Experimental Settings}} We train our INS framework on NeRF-Synthetic~\cite{nerf} dataset and Local Light Field Fusion(LLFF)~\cite{mildenhall2019local} dataset. NeRF-Synthetic consists of complex scenes with 360-degree views, where each scene has a central object with 100 inward-facing cameras distributed randomly on the upper hemisphere. Both rendered images and ground truth meshes are provided in NeRF-Synthetic dataset. LLFF dataset consists of forward-facing scenes, with fewer images. We implement INS on the same architecture and training strategy with the original NeRF~\cite{nerf}. $\lambda_1$, $\lambda_2$ and $\lambda_3$ are set as zero in the first 150k iterations and then set as 1e6, 1 and 1e8 in the following 50k iterations. The self-distilled density supervision depicted in Figure~\ref{fig:arc_nerf} is generated from the CIM with 150k iterations pre-training. Adam optimizer is adopted with learning rates of 0.0005.
Hyper-parameters are carefully tuned via grid searches and the best configuration is applied to all experiments. 
All experiments are trained on one NVIDIA RTX A6000 GPU. We retrain Style3D~\cite{chiang2022stylizing} in NeRF-Synthetic and LLFF datasets using their provided code and setting. We train all methods using the same number of style images for fair comparisons.

\paragraph{\textbf{Results}}
In Figure~\ref{fig:sole_exp_on_nerf}, we can see INS generates faithful and view-consistent results for new viewpoints, with rich textures across scenes and styles.
We further compare INS with three state-of-the art methods, including 3D neural stylization~\cite{chiang2022stylizing}, image-based stylization methods~\cite{huang2017arbitrary,johnson2016perceptual}.
As is shown in Figure~\ref{fig:compare}, We can see that stylizations from image-based methods produce noisy and view-inconsistent stylization as they transfer styles based on a single image. Style3D~\cite{chiang2022stylizing} generates blur results as it still relies on convolution networks (a.k.a. hypernetwork) to generate the MLP weights for the subsequent volume rendering.
Our proposed implicit neural stylization method is trained to preserve correct scene geometry as well as capture global context, generating better view-consistent stylizations.
\begin{figure*}
\centering
\begin{tabular}{c c c c c c}
    \includegraphics[width=0.16\textwidth]{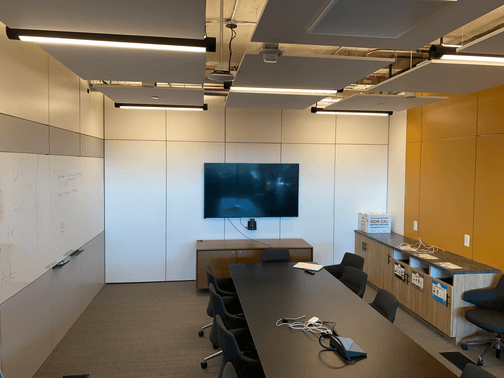} &
      \includegraphics[height=0.12\textwidth]{figures/the_scream.jpg} &
  \includegraphics[width=0.16\textwidth]{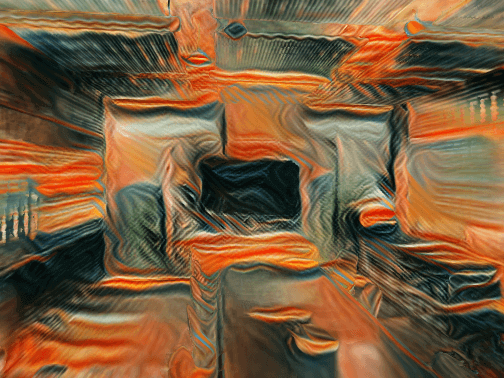} & 
  \includegraphics[width=0.16\textwidth]{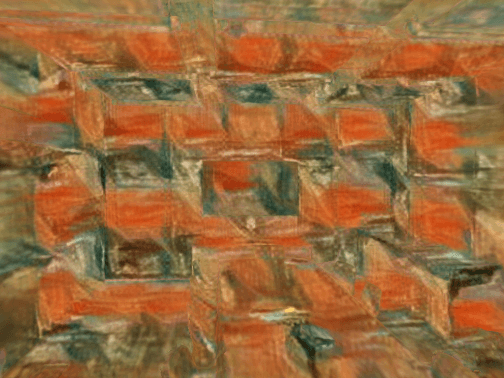} &
  \includegraphics[width=0.16\textwidth]{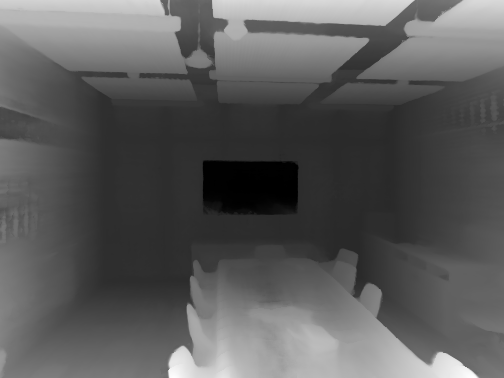} &
  \includegraphics[width=0.16\textwidth]{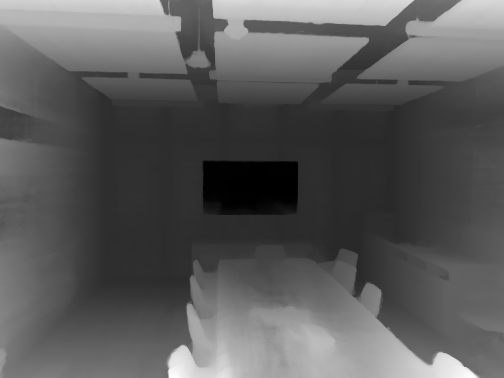}\\
  Color Image &Style &Color w/ SS &Color w/o SS &Depth w/ SS &Depth w/o SS   \\
\end{tabular}
\caption{Qualitative evaluation on the effectiveness of Sampling Stride technique (SS). We can see INS with a larger receptive field generate more satisfying stylization.}\label{fig:patch_stride_abalation}
\end{figure*}
\subsection{Stylization on Signed Distance Function}
\begin{figure*}[t]
\centering
\resizebox{0.9\textwidth}{!}{
\begin{tabular}{c c c c }
    \includegraphics[width=0.22\textwidth]{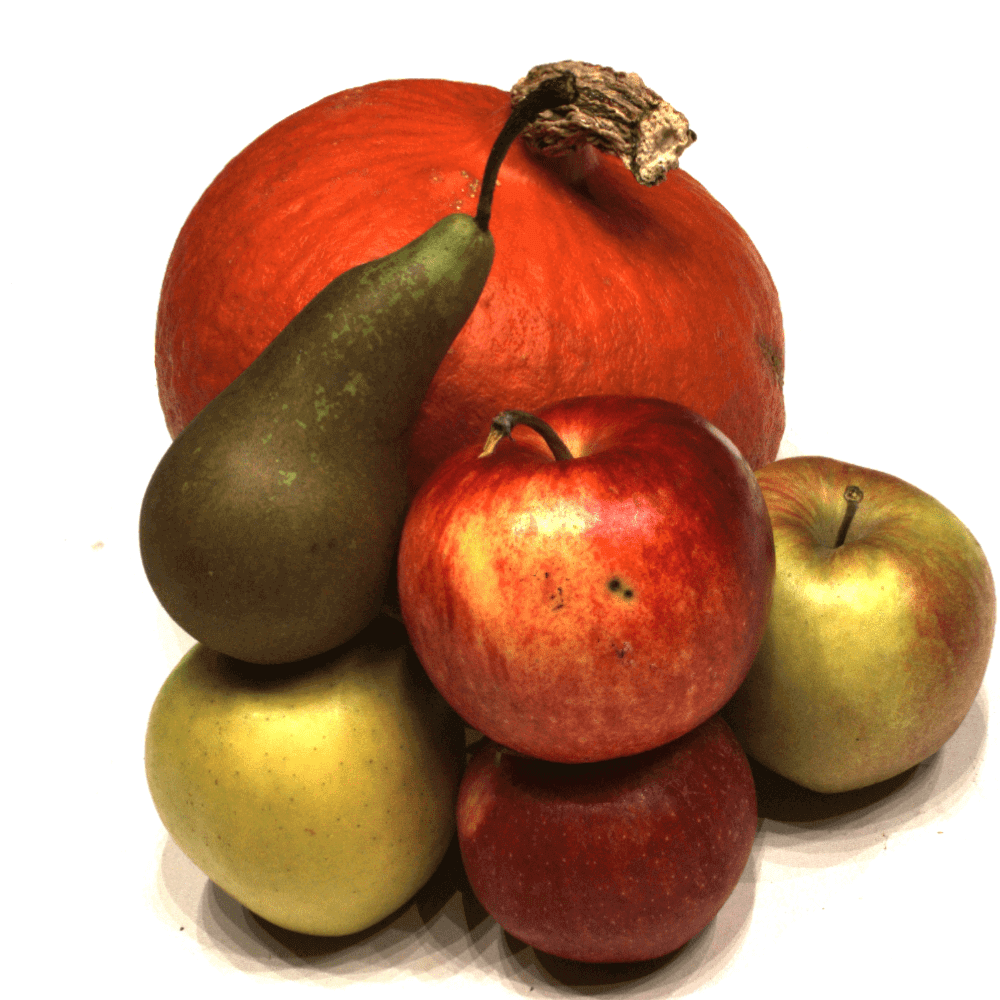} &
   \includegraphics[width=0.22\textwidth]{figures/face.jpg} & 
   \includegraphics[width=0.22\textwidth]{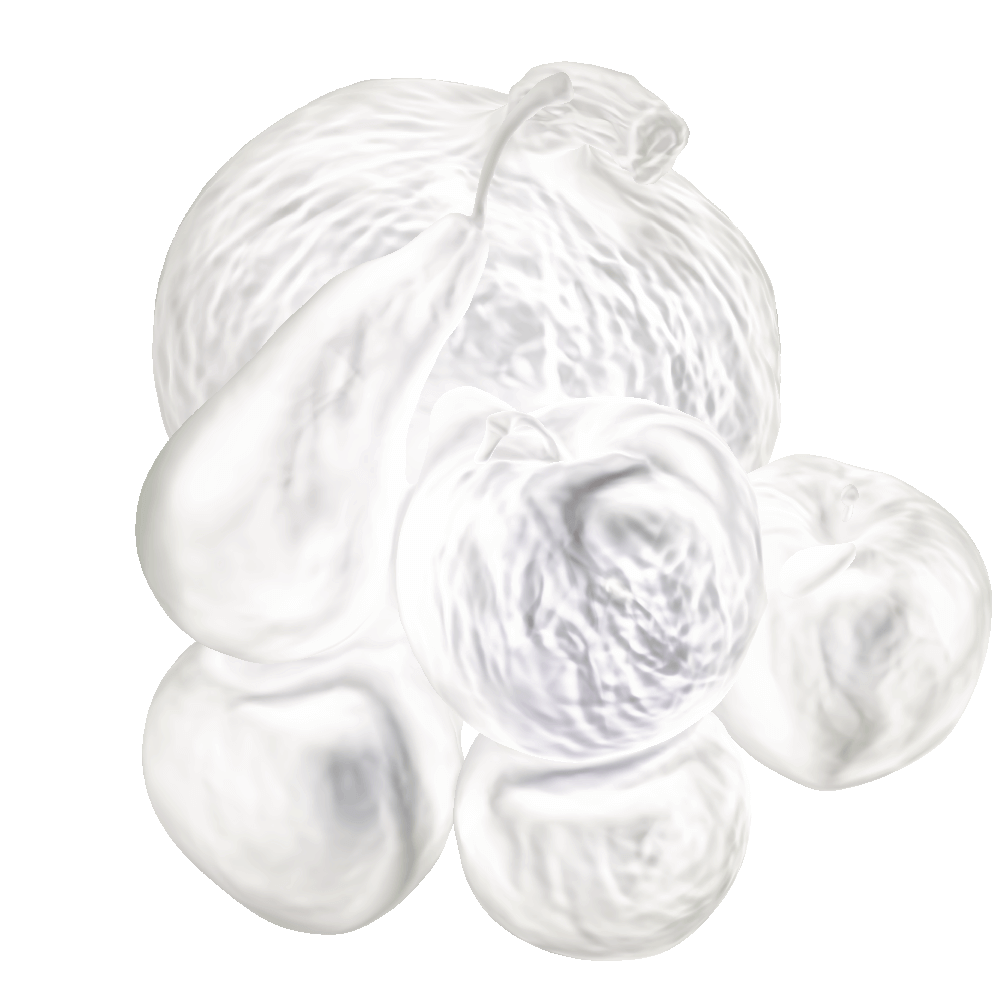} &
   \includegraphics[width=0.22\textwidth]{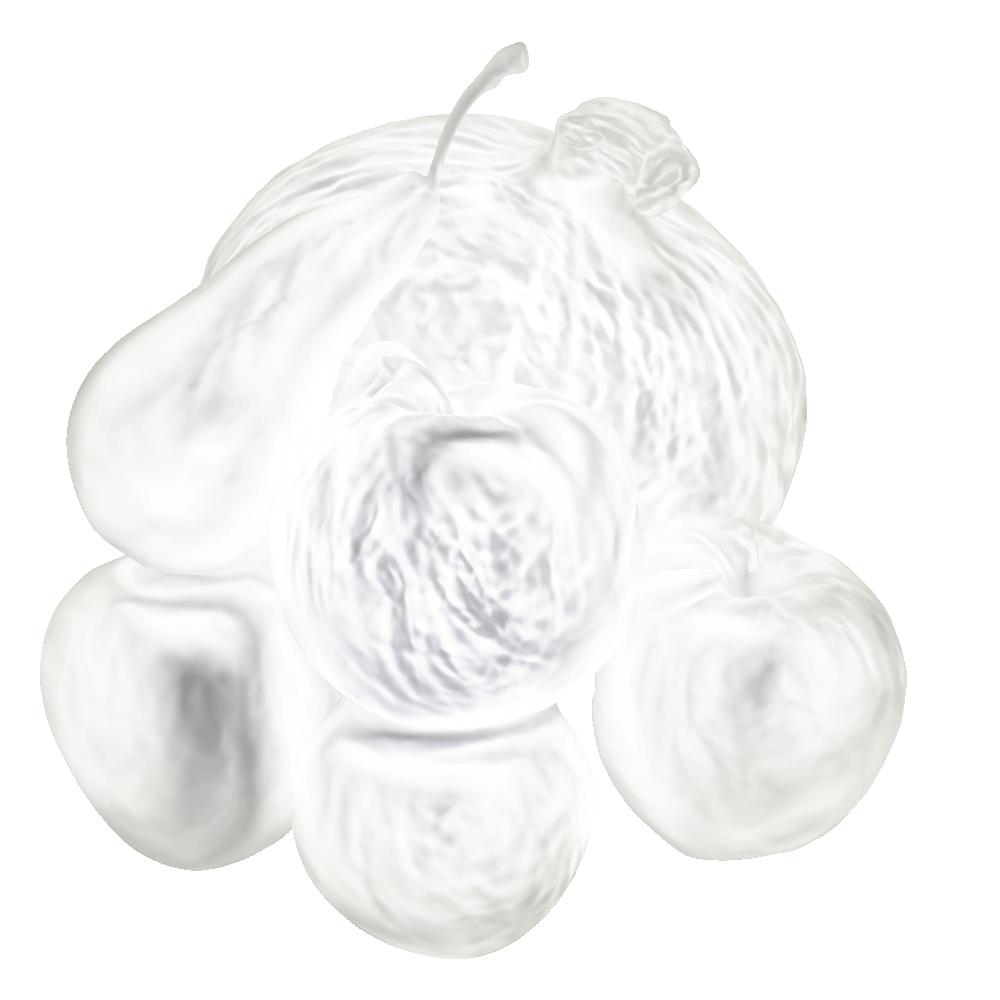}\\
   Color Image &Style Image & SDF(view1) &SDF(view2)   \\
   \includegraphics[width=0.22\textwidth]{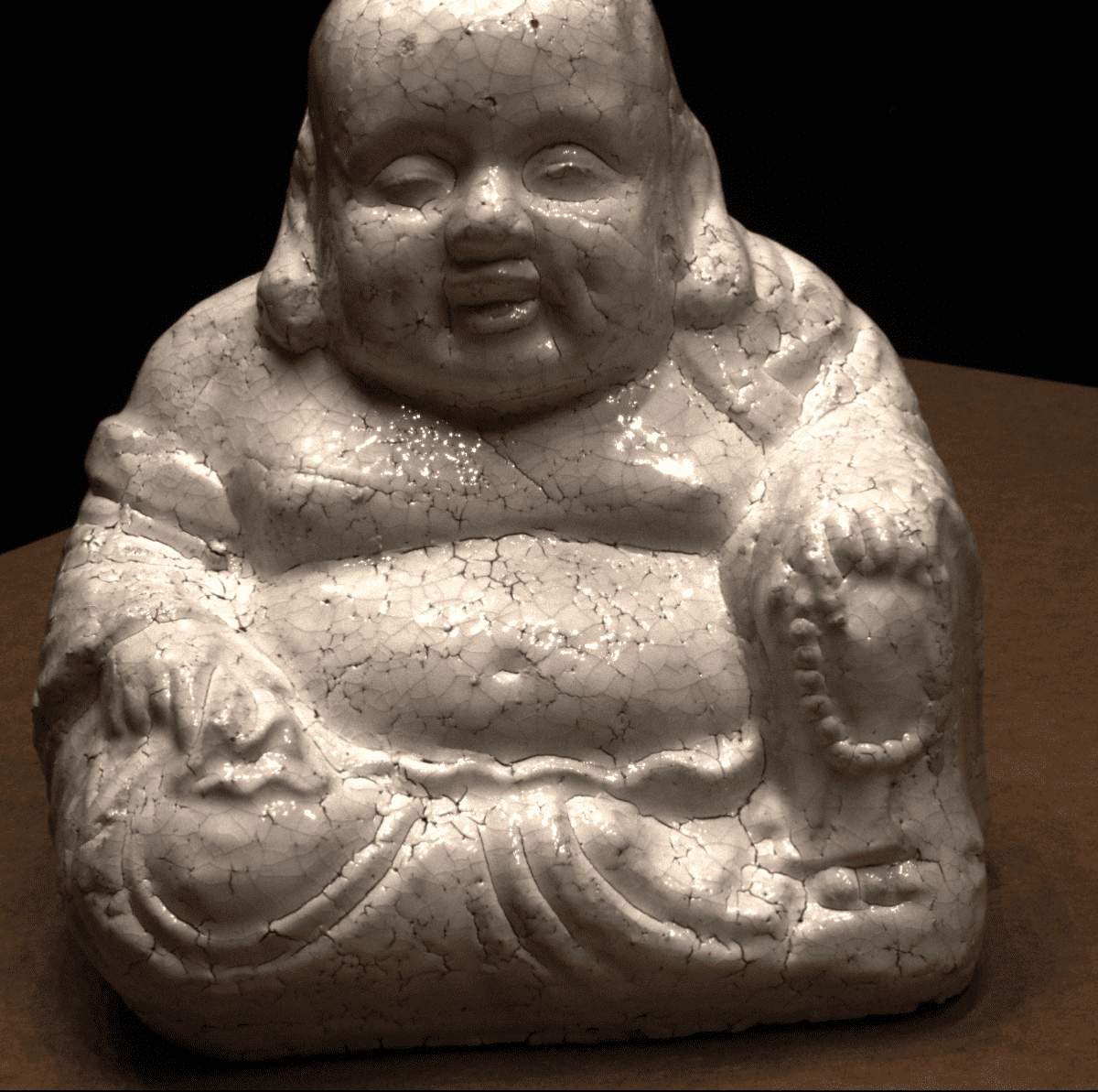} &
   \includegraphics[height=0.22\textwidth]{figures/the_scream.jpg} & 
   \includegraphics[width=0.22\textwidth]{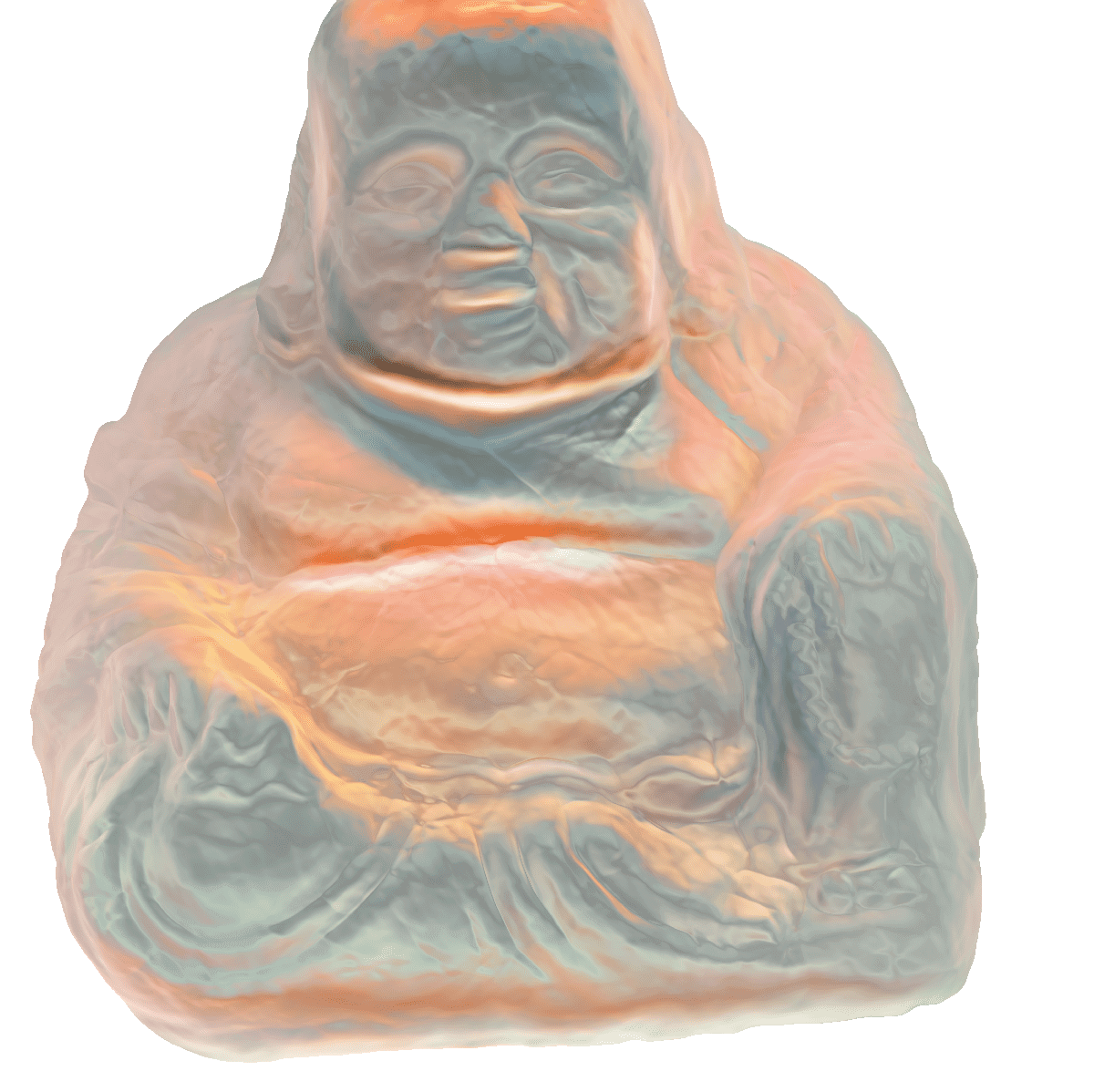} &
   \includegraphics[width=0.22\textwidth]{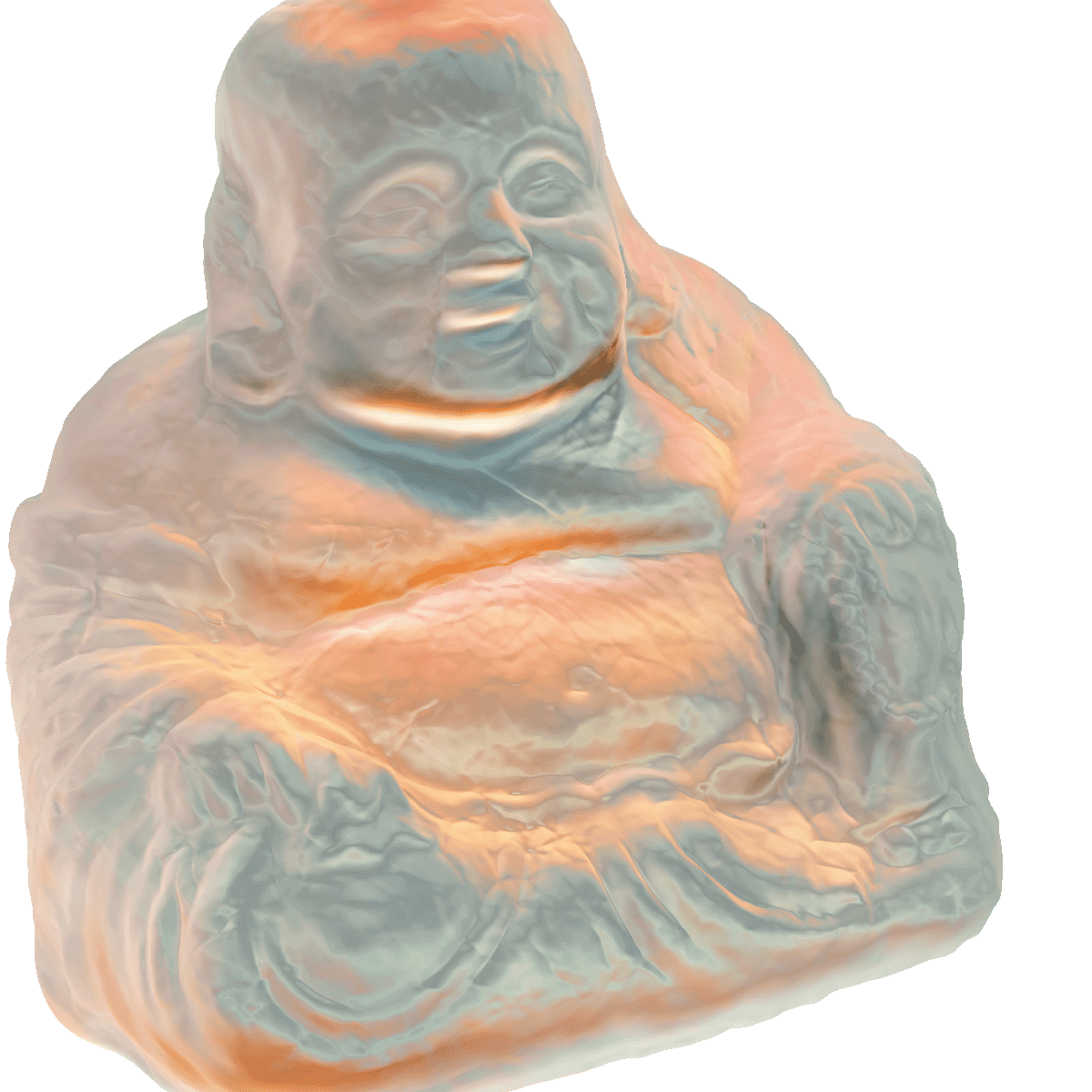}\\
    Color Image &Style Image & SDF(view1) &SDF(view2)   \\
   \end{tabular}}
\caption{\textbf{Visualization results of applying INS to the Signed Distance Function.} Given multi-view color images and style images, the proposed INS can learn the style statistics for the disentangled geometry and appearance.}\label{fig:res_on_idr}
\end{figure*}

\paragraph{\textbf{Experimental Settings}}
DeepSDF~\cite{park2019deepsdf} only learns the 3D geometry from given inputs. Later work IDR~\cite{yariv2020multiview} extend it to reconstruct both 3D surface and appearance. We follow IDR~\cite{yariv2020multiview} to implement the implicit neural stylization framework.
To encode style statistic onto IDR, we project the learned textured SDF into multi-view images and implement our style loss on the rendered results. 
In the experiments, we picked 2 scenes from the DTU dataset~\cite{aanaes2016large}, where each scene consists of 50 to 100 images and object masks captured from different angles.
Similar to NeRF, we pre-train the IDR model for chosen scenes by minimizing the loss between the ground truth image and the rendered result. Then both the SDF network and rendering network are jointly optimized the proposed framework from projected views.
Note that due to IDR's architectural design, we are no longer able to impose self-distilled geometry consistency loss. Instead, we employ content/style loss in the masked region, which is similar to~\cite{yariv2020multiview}.
Besides, we observe that the SDF representation is more sensitive to parameter variations. To maintain intact geometries, we adjust the learning rate for the SDF network to $10^{-11}$ times smaller than the rendering network. 
\paragraph{\textbf{Results}}
As are shown in Figure~\ref{fig:res_on_idr}, the visualizations of two-view SDF representation demonstrate that both the learned appearance and geometry have deformed to fit the given style statistics.

\subsection{Conditional Style Interpolation}
\begin{figure*}[t!]
\centering
\begin{tabular}{c c c c c c}
   \includegraphics[width=0.14\textwidth]{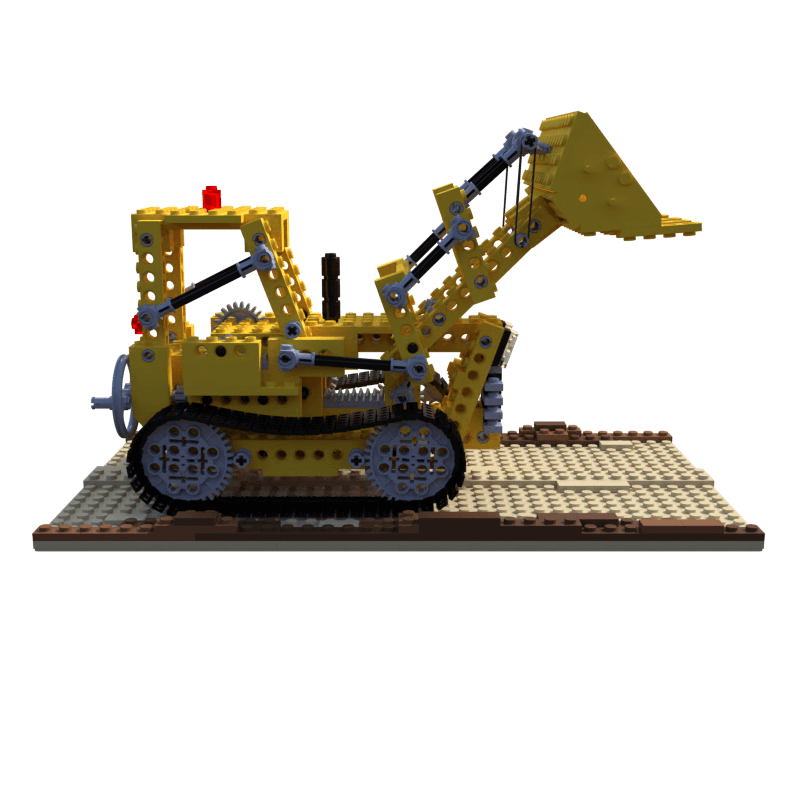} &
      \includegraphics[width=0.14\textwidth]{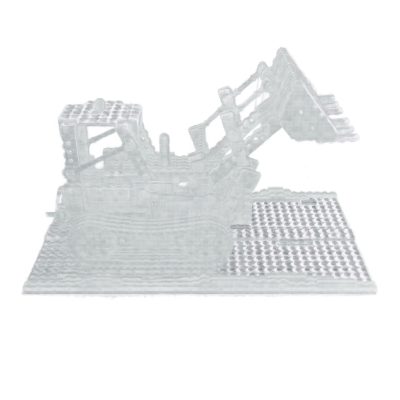} &
  \includegraphics[width=0.14\textwidth]{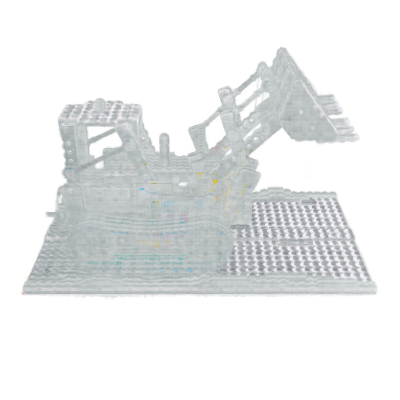} & 
   \includegraphics[width=0.14\textwidth]{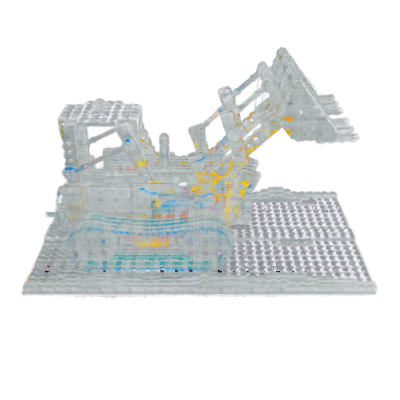} & 
   \includegraphics[width=0.14\textwidth]{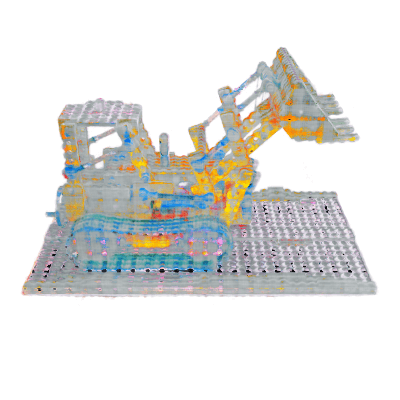} &
    \includegraphics[width=0.14\textwidth]{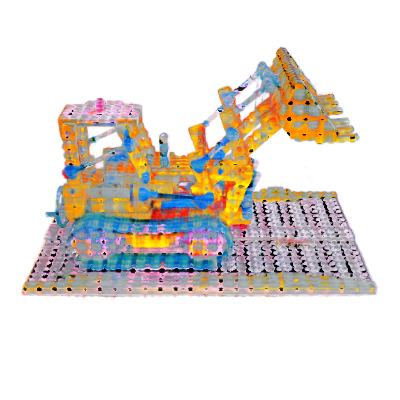}\\
   \scriptsize{Color Image}  
   &\scriptsize{[1, 0, ..., 0]}  
   &\scriptsize{[0.9, 0.1, ..., 0]}  
    &\scriptsize{[0.8, 0.2, ..., 0]} 
    &\scriptsize{[0.7, 0.3, ..., 0]}
    &\scriptsize{[0.6, 0.4, ..., 0]}\\
    \includegraphics[width=0.10\textwidth]{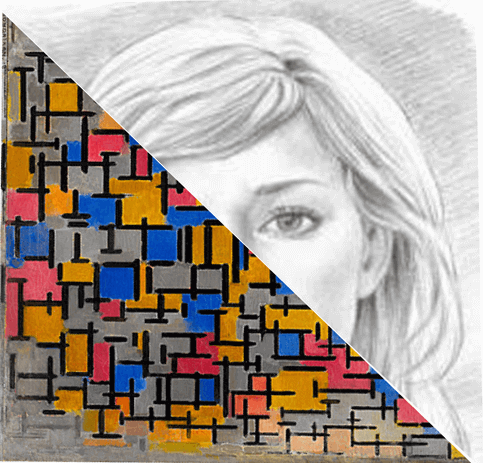} &
    \includegraphics[width=0.14\textwidth]{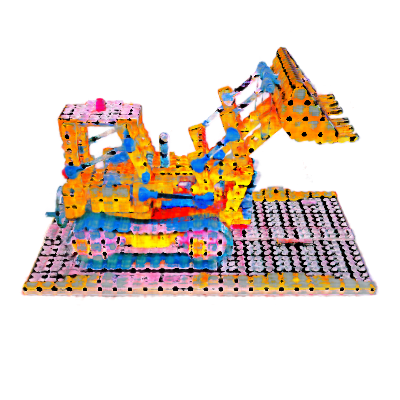} &
   \includegraphics[width=0.14\textwidth]{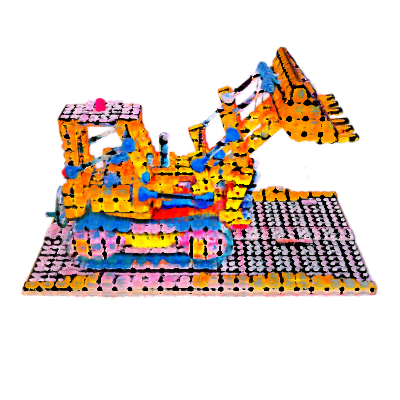}  & 
   \includegraphics[width=0.14\textwidth]{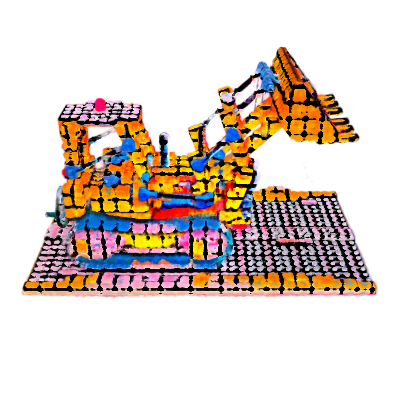} &
   \includegraphics[width=0.14\textwidth]{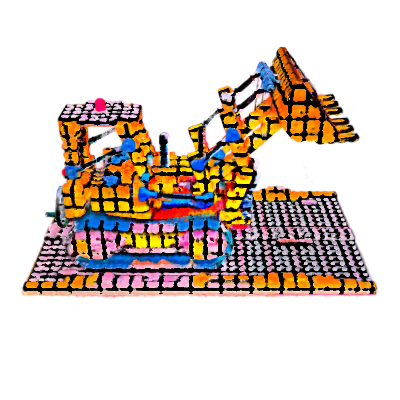}  &
    \includegraphics[width=0.14\textwidth]{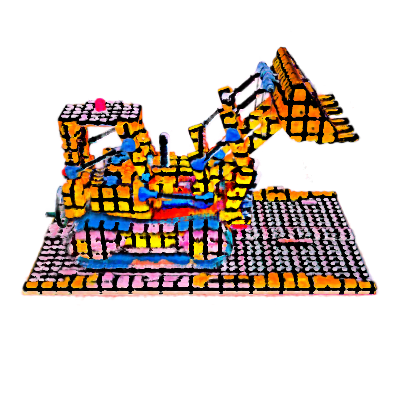}  
\\
   \scriptsize{Style Images}  
   &\scriptsize{[0.5, 0.5, ..., 0]}
   &\scriptsize{[0.4, 0.6, ..., 0]}
   &\scriptsize{[0.3, 0.7, ..., 0]}
   &\scriptsize{[0.2, 0.8, ..., 0]} 
   &\scriptsize{[0, 1, ..., 0]}
   \\
\end{tabular}
\caption{Visualization of the mixture of two styles using different mixture weights. We can see the style smoothly transfers from one style to another.
}\label{fig:supp_conditional}
\end{figure*}

Training with style-conditioned one-hot embedding, we can interpolate between style images to mix multiple styles with arbitrary weights. Specifically, we train INS on NeRF with two style images along with a two-dimensional one-hot vector as conditional code. After training, we mix the two style statistics by using a weighted two-dimensional vector. As is shown in Figure~\ref{fig:supp_conditional}, the synthesized results can smoothly transfer from the first style to the second style when we linearly mix the two style embeddings at inference time.

\begin{figure*}[t!]
\centering
\begin{tabular}{c c c c c}
    \includegraphics[width=0.17\textwidth]{figures/r_19.png} &
   \includegraphics[width=0.17\textwidth]{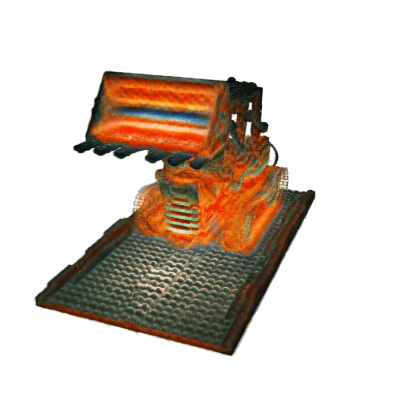} & 
   \includegraphics[width=0.17\textwidth]{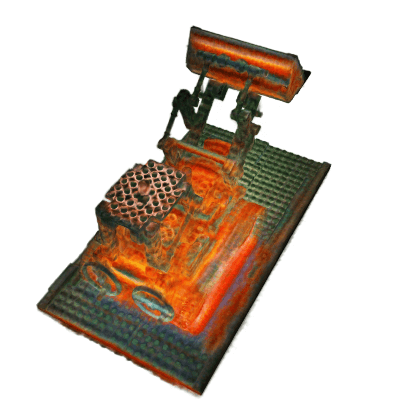} &
   \includegraphics[width=0.17\textwidth]{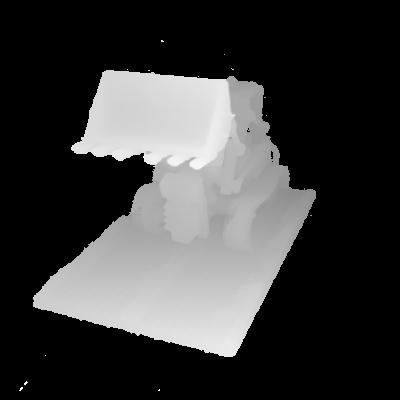} &
   \includegraphics[width=0.17\textwidth]{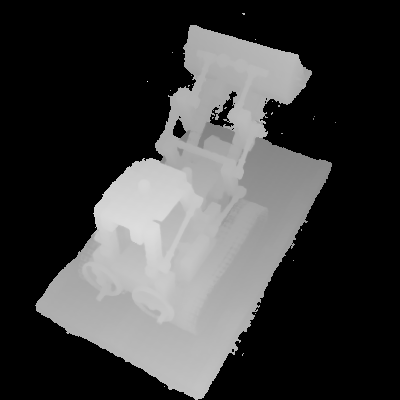}\\
   Color Image &Color w/ GC &Color w/ GC &Depth w/ GC &Depth w/ GC   \\
   \includegraphics[height=0.17\textwidth]{figures/the_scream.jpg} &
   \includegraphics[width=0.17\textwidth]{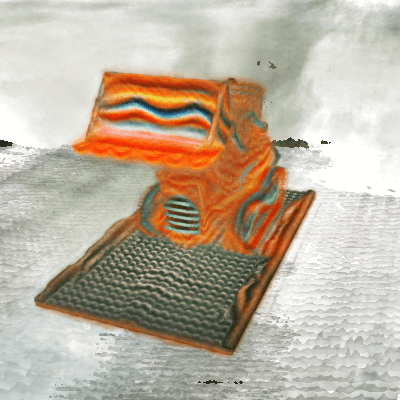} & 
   \includegraphics[width=0.17\textwidth]{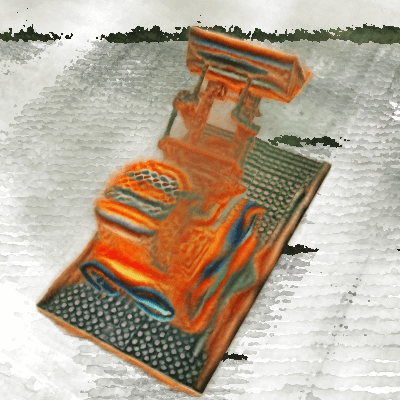} &
   \includegraphics[width=0.17\textwidth]{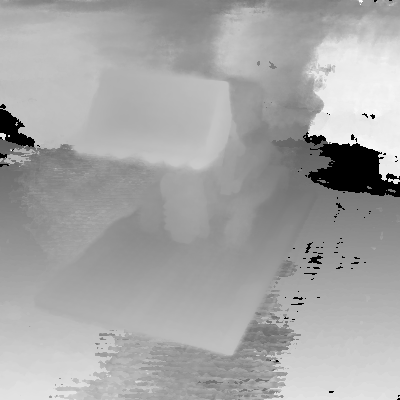} &
   \includegraphics[width=0.17\textwidth]{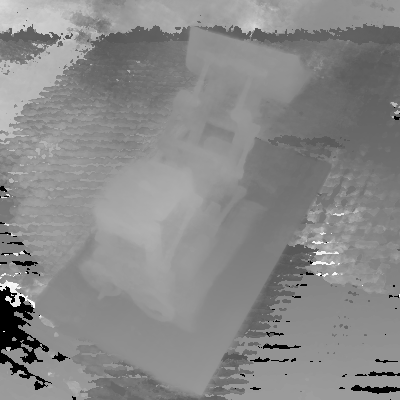}\\
   Style Image &Color w/o GC &Color w/o GC &Depth w/o GC &Depth w/o GC   \\
     
\end{tabular}
\caption{Qualitative results of the self-distilled geometry consistency(``GC''). Both the rendered images and depth maps are shown for validation.}\label{fig:densityloss}
\end{figure*}

\subsection{Ablation Study}

\begin{figure*}[!t]
\centering
\includegraphics[width=0.95\textwidth]{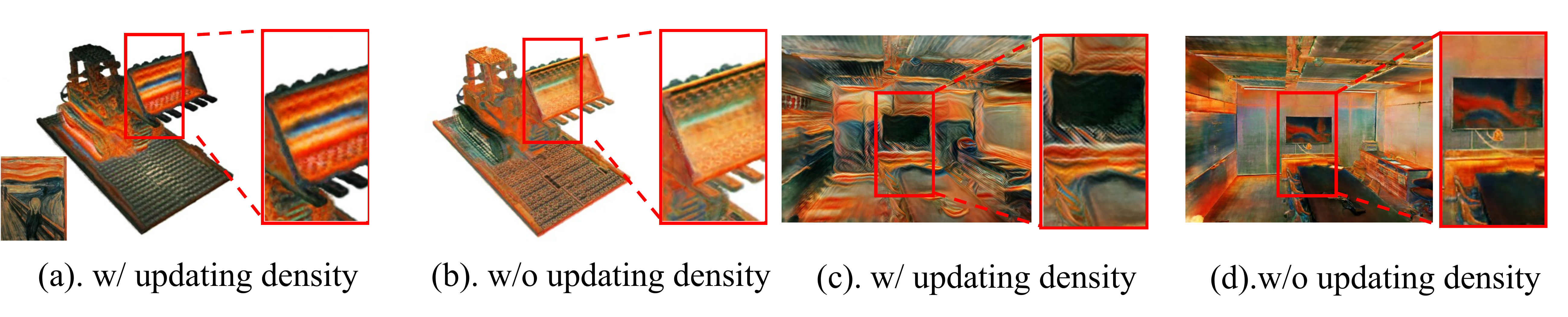}
\caption{Ablation study of optimizing style-conditioned density and color. By updating the density, it can render richer textures than only updating the color branch.}
\label{fig:without_density}
\end{figure*}

\paragraph{\textbf{Effect of Updating NeRF's Geometry Field}} The geometry modification in the density branch enables a more flexible stylization, by stylizing shape tweaks on the object surface. As shown in Figure.~\ref{fig:without_density}, only updating the color branch easily collapses to the low-level color transformation instead of the painterly texture transformation, violating our original goal. This phenomenon is also addressed in the previous 3D mesh stylization~\cite{kato2018neural}, where they explicitly model the 3D shape by deforming the mesh vertexes. 

\paragraph{\textbf{Effect of Self-distilled Geometry Consistency}} To evaluate the effectiveness of the proposed geometry consistency regularizer, we visualize the front and back viewpoints of the synthesized color images and depth maps. As is shown in Figure~\ref{fig:densityloss}, the proposed self-distilled geometry consistency learns a good trade-off between stylization and clean geometry.

\paragraph{\textbf{Should INS Learned with Larger Receptive Field?}} To investigate the effect by using the Sampling Stride (SS) Ray Sampling strategy, we conduct a comparison of ray sampling with and without sampling stride for NeRF stylization. For a fair comparison, we set the ray number as 64$\times$64 in both settings. INS with sampling stride covers the content resolution of (64$\times$s) $\times$ (64$\times$s) where $s$ indicates sampling strides depicted in Section~\ref{sec_Sampling stride Ray Sampling} and here we set $s$=4. Figure~\ref{fig:patch_stride_abalation} shows that INS with SS achieves significantly better visual results, as it results in a higher receptive field in perceiving content statistics.

\section{Conclusions}
In this work, we present a Unified Implicit Neural Stylization framework (INS) to stylize complex 2D/3D scenes using implicit function. We conduct a pilot study on different types of implicit representations, including 2D coordinate-based mapping function, Neural Radiance Field, and Signed Distance Function.
Comprehensive experiments demonstrate that the proposed method yields photo-realistic images/videos with visually consistent stylized textures. One limitation of our work lies in the training efficiency issue, similar to most implicit representation, rendering a style scene requires several hours of training, precluding on-device training. Addressing this issue could become a future direction.

\bibliographystyle{splncs04}
\bibliography{egbib}
\end{document}